\documentclass[letterpaper]{article} 
\usepackage[preprint]{aaai2027}   

\usepackage[hyphens]{url}  
\usepackage{graphicx} 
\usepackage{natbib}  
\usepackage{caption} 
\usepackage{algorithm}
\usepackage{algorithmic}

\usepackage{newfloat}
\usepackage{listings}
\DeclareCaptionStyle{ruled}{labelfont=normalfont,labelsep=colon,strut=off} 
\floatstyle{ruled}
\newfloat{listing}{tb}{lst}{}
\floatname{listing}{Listing}

\usepackage{booktabs}

\usepackage[hyphens]{url}  
\usepackage{graphicx} 
\usepackage{natbib}  
\usepackage{caption} 
\usepackage{algorithm}
\usepackage{algorithmic}
\usepackage{newfloat}
\usepackage{listings}
\usepackage{booktabs}
\usepackage{array}
\usepackage{xcolor}

\DeclareCaptionStyle{ruled}{labelfont=normalfont,labelsep=colon,strut=off} 
\floatstyle{ruled}
\newfloat{listing}{tb}{lst}{}
\floatname{listing}{Listing}

\title{Memory Is Not Always Needed:\\Characterizing Conditional Memory in Scientific Reasoning}
\author{
    Written by AAAI Press Staff\textsuperscript{\rm 1}\thanks{With help from the AAAI Publications Committee.}\\
    AAAI Style Contributions by Peter Patel Schneider,
    Sunil Issar,\\
    J. Scott Penberthy,
    George Ferguson,
    Hans Guesgen,
    Francisco Cruz\equalcontrib\corresponding,
    Marc Pujol-Gonzalez\equalcontrib\corresponding
}
\affiliations{
    \textsuperscript{\rm 1}Association for the Advancement of Artificial Intelligence\\

    1101 Pennsylvania Ave, NW Suite 300\\
    Washington, DC 20004 USA\\
    proceedings-questions@aaai.org
}

\title{Memory Is Not Always Needed: Characterizing Conditional Memory in \\ Scientific Reasoning}

\author{
    Zhen Bi\textsuperscript{\rm 1}\thanks{Equal Contribution},
    Xueshu Chen\textsuperscript{\rm 1}\footnotemark[1],
    Yan Wang\textsuperscript{\rm 2},
    Zhizhi Peng\textsuperscript{\rm 1},
    Haosen Hong\textsuperscript{\rm 3},
    \\
    Zhen Wang\textsuperscript{\rm 1},
    Zhixuan Chu\textsuperscript{\rm 4},
    Bingyu Zhu\textsuperscript{\rm 2},
    Jungang Lou\textsuperscript{\rm 1}\thanks{Corresponding Author.}
}

\affiliations {
    \textsuperscript{\rm 1}Huzhou Normal University, \textsuperscript{\rm 2}Alibaba Group\\
    \textsuperscript{\rm 3}Bota Biosciences,
    \textsuperscript{\rm 4}Zhejiang University\\
    bizhen\_zju@zju.edu.cn
}

\begin{document}

\maketitle

\begin{abstract}
{
Scientific reasoning requires language models to retrieve specialized knowledge and incorporate it reliably into multi-step computation. 
Conditional memory provides an explicit lookup pathway that complements dense neural representations, but its usefulness is inherently input- and computation-dependent: retrieved information may repair missing scientific associations, yet it may also introduce distracting shortcuts or interfere with reasoning that the base model can already perform correctly. 
In this work, we systematically investigate when, where, and to what extent conditional memory should participate in scientific reasoning. 
We characterize the scientific knowledge boundary and controlled interventions on memory-enabled knowledge-circuit nodes.
Based on these analyses, we propose a Knowledge Boundary-Aware Router that uses task-specific input proxies available before generation to determine whether memory is activated, which layer–stage nodes receive memory signals, and how strongly these signals contribute.
Experiments on biological and chemical reasoning benchmarks, covering two backbone families and six task types, show that memory effects vary substantially across inputs, tasks, and injection locations. 
Compared with static and activation-rate-matched random routing, our approach more consistently preserves beneficial memory contributions while suppressing memory-induced regressions, establishing selective memory allocation as an important principle for reliable scientific reasoning.
}
\end{abstract}

\section{Introduction}
{
Scientific discovery increasingly requires models to reason over complex biological, chemical, physical, and mathematical systems \cite{DBLP:journals/natmi/JablonkaSOS24,DBLP:journals/natmi/RossBCPMD22,DBLP:journals/nature/MerchantBSACC23,DBLP:journals/nature/BoikoMKG23,DBLP:journals/nature/RomeraParedesBNBKDREWFKF24}. 
Improving the scientific reasoning capabilities of large language models is therefore essential for reliable scientific analysis and discovery \cite{WeiEtAl2025AgenticScience}. 
At its core, this requires models not only to access specialized knowledge, but also to integrate it faithfully throughout multi-step inference.
}

\begin{figure}[h]
    \centering
    \includegraphics[width=0.45\textwidth]{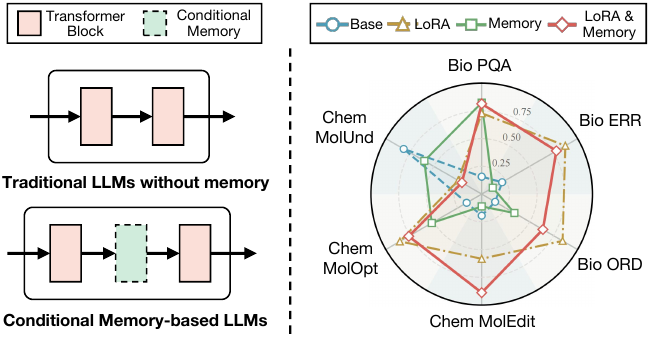}
    \caption{
    {
    \textbf{Left:} Conditional memory is a model-internal lookup-and-injection mechanism: memory modules are placed within Transformer computation.
    \textbf{Right:} Even on knowledge-intensive scientific reasoning tasks, memory produces mixed effects rather than uniform gains, motivating selective activation.
    }    
    }
    \label{fig:intro-placeholder}
\end{figure}

{
Existing approaches mainly rely on domain-specialized models or retrieval-augmented generation \cite{DBLP:conf/nips/ZhangHZDYWYD024,DBLP:conf/iclr/WenTDD0XT24, DBLP:journals/corr/abs-2508-15763,DBLP:conf/icml/GuuLTPC20}.
Although they store knowledge differently, both lack a clear separation between knowledge lookup and neural computation. Conditional memory offers a complementary lookup path \cite{DBLP:conf/acl/ChengZDCWXHYHZL26} and is particularly promising for knowledge-intensive scientific reasoning, where specialized entities, motifs, structures, and rules recur frequently.
}

{
Recent studies have applied conditional memory to genomic foundation models and molecular language models, reporting benefits from the explicit lookup of biological motifs and molecular patterns \cite{DBLP:journals/corr/abs-2601-22203,DBLP:journals/corr/abs-2606-12113}.
However, these results remain concentrated in specific scientific settings and do not establish that memory should be activated uniformly in broader scientific reasoning. 
Across heterogeneous tasks, memory may repair a missing scientific association in one case, provide little benefit in another, or introduce misleading shortcuts that disrupt an otherwise correct reasoning path (shown in Figure~1). 
Its utility must therefore be evaluated relative to the knowledge boundary of the memory-free base model and the computational location at which memory is injected.
}

{
Therefore, in this work, we investigate \textbf{when conditional memory is a viable computational mechanism for scientific reasoning} and we systematically study its effects through the scientific knowledge boundary. 
Specifically, through behavioral knowledge-boundary analysis and controlled layer–stage interventions, we show that memory utility depends jointly on the input, injection location, and contribution strength. 
These findings lead to a \textbf{Knowledge Boundary-Aware Router} that determines whether, where, and how strongly memory participates, revealing selective memory allocation as a broader design principle for scientific reasoning models.
}

Our contributions are threefold:
\begin{itemize}
    \item 
    {
        We conduct a systematic empirical study of conditional memory in scientific reasoning across biological and chemical domains, two backbone families, and six task types. 
    }
    {
        We show that memory is a viable but inherently non-uniform augmentation: it repairs some scientific reasoning failures while inducing regressions in others.
    }
    \item
    {
        We characterize this heterogeneity through behavioral capability-boundary indicators and layer–stage memory interventions, revealing that memory utility depends on both the input regime and its location in the reasoning computation.
    }
    \item
    {
        We translate these findings into a  boundary-aware routing mechanism that selectively activates memory using task-specific input proxies. The router outperforms activation-rate-matched random routing and avoids several regressions caused by static memory configurations.
    }
\end{itemize}

\begin{figure*}[t]
    \centering
    \includegraphics[width=0.98\textwidth]{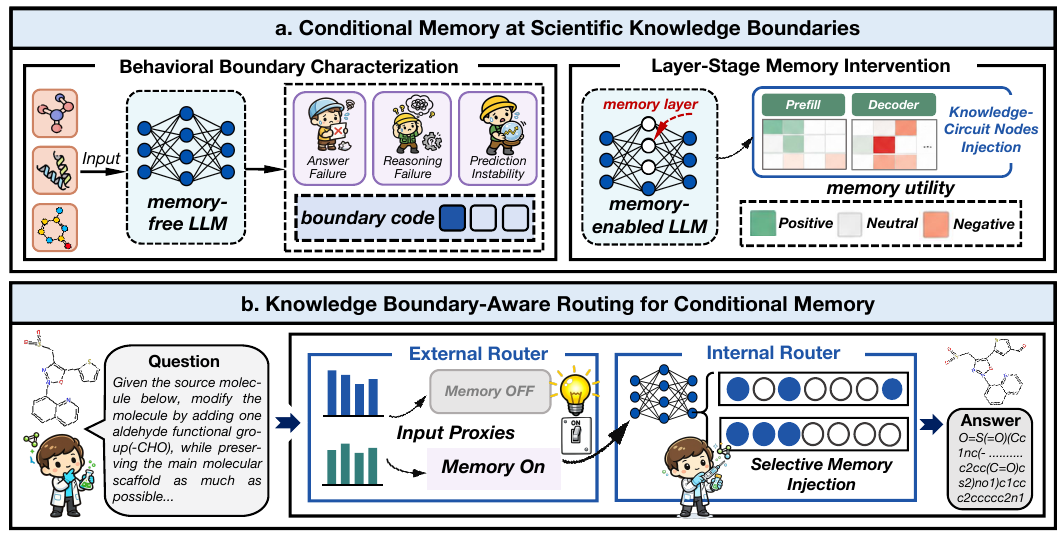}
    \caption{Overview of our work.
    We analyze the problem from two complementary perspectives (\textbf{the behavioral and knowledge-circuit view}).
    Then we propose the \textbf{Knowledge Boundary-Aware Routing for
    Conditional Memory}( \textbf{the external boundary-aware data router} and \textbf{the internal boundary-aware parameter router}).
    }
    \label{fig:main-placeholder}
\end{figure*}

\section{Related Work}

\subsection{Conditional Memory-Based LLMs}
{
Memory-augmented LLMs supplement dense parameters with selectively accessed stores. 
Sparse key-value layers expand capacity \cite{DBLP:conf/nips/LampleSRDJ19} and lookup-based conditional memory, instantiated by Engram, uses suffix $n$-gram lookup as a complementary sparsity axis \cite{DBLP:conf/acl/ChengZDCWXHYHZL26}.
Memory-expert architectures inject pre-stored knowledge through token-level memory experts \cite{ding2026meki} and datastore methods retrieve targets, passages or neighboring chunks \cite{DBLP:conf/iclr/KhandelwalLJZL20,DBLP:conf/nips/LewisPPPKGKLYR020,DBLP:conf/icml/BorgeaudMHCRM0L22}.
Activation-based memories reuse cached historical states \cite{DBLP:conf/iclr/WuRHS22,DBLP:conf/nips/Wang0CLYGW23}.
Writable systems further support dialogue recall, episodic or post-deployment updates, and scalable sparse memory \cite{DBLP:journals/corr/abs-2312-17257,DBLP:journals/corr/abs-2403-11901,DBLP:conf/icml/WangGCJLYYLLYSM24,DBLP:conf/icml/BergesOHYZG25}. 
Existing work emphasizes aggregate gains and we study when memory is needed and causally useful at scientific boundaries.
}

\subsection{Scientific Reasoning Knowledge Boundaries}
{
Scientific reasoning benchmarks span broad STEM knowledge, specialized problem solving, and theorem application \cite{DBLP:conf/iclr/HendrycksBBZMSS21,DBLP:conf/icml/WangHL0ZSLZS024,DBLP:journals/corr/abs-2311-12022,DBLP:conf/emnlp/ChenYKLWMXWX23}. 
Domain-specific models, benchmarks, and reviews further cover Chinese biomedical language understanding \cite{ZhangEtAl2022CBLUE}, protein representation learning with Gene Ontology knowledge \cite{ZhangEtAl2022OntoProtein}, ocean-science tasks \cite{BiEtAl2024OceanGPT}, and healthcare LLM techniques and applications \cite{HuEtAl2025HealthcareLLM}. 
Final-answer correctness, however, cannot distinguish knowledge gaps from faulty derivation. Process-level feedback and reward models expose intermediate errors \cite{DBLP:journals/corr/abs-2211-14275,DBLP:conf/iclr/LightmanKBEBLLS24}. Uncertainty studies probe self-evaluation and elicited confidence \cite{DBLP:journals/corr/abs-2207-05221,DBLP:conf/iclr/XiongHLLFHH24}, while sampled-response disagreement and semantic entropy reveal repeated-inference instability \cite{DBLP:conf/emnlp/ManakulLG23,DBLP:journals/nature/FarquharKKG24}.
We synthesize these signals into final-answer failure, reasoning-step failure, and repeated-inference instability, which characterize the base model's scientific reasoning boundary for conditional-memory analysis.
}

\section{Conditional Memory at Scientific Knowledge Boundaries}

{
In this section, we analyze the problem from two complementary perspectives. The behavioral view identifies \emph{when} an input reaches the base model's knowledge boundary:
final-answer failure, reasoning-step failure, and prediction instability expose incorrect outcomes, faulty derivations, and unreliable repeated inference. The
knowledge-circuit view identifies \emph{where} memory can affect computation: a hidden state queries the store, whose retrieved value is injected through a native gate at a selected layer and stage. 
}

\subsection{Behavioral Scientific Knowledge Boundary Characterization}

{
We assess memory against the memory-free base model, since it may repair
missing associations or disrupt otherwise sufficient reasoning.
Memory utility varies by input.
}
For each example $x_i$, we define a behavioral boundary code with respect to the memory-free backbone $M_{\mathrm{base}}$. Given a task-level answer evaluator $E_{\mathrm{ans}}$, a process-level evaluator $E_{\mathrm{step}}$, and repeated-inference consistency score $C_i$, we compute
\begin{equation}
b_i^{A}=\mathbf{1}\left[E_{\mathrm{ans}}(M_{\mathrm{base}}(x_i), y_i)<\tau_A\right],
\end{equation}
\begin{equation}
b_i^{R}=\mathbf{1}\left[E_{\mathrm{step}}(M_{\mathrm{base}}(x_i), y_i)<\tau_R\right],
\end{equation}
\begin{equation}
b_i^{P}=\mathbf{1}\left[C_i<\tau_P\right].
\end{equation}
Here $b_i^{A}$, $b_i^{R}$, and $b_i^{P}$ mark final-answer failure, reasoning-step failure, and prediction instability, respectively. 
We encode them as
\begin{equation}
\mathbf{q}_i = \left(b_i^{A}, b_i^{R}, b_i^{P}\right) \in \{0,1\}^{3}.
\end{equation}

{
The indicators are nonexclusive: an input may have a correct final answer despite faulty derivation, be stably incorrect, or be unstable despite one correct sample. The code records these distinct boundary behaviors rather than imposing a single difficulty ranking, allowing us to test whether memory repairs a failure regime or merely changes aggregate performance.
}

{
In Eq.~(4), $\mathbf{q}_i=\mathbf{0}$ denotes a non-boundary example, while a nonzero entry identifies the corresponding boundary type. Computing $\mathbf{q}_i$ can require labels or repeated inference, so we use it only offline to characterize samples and calibrate the router. At inference, Section~4.1 instead uses task-specific input proxies available before generation.
}

\providecommand{\stdcell}[2]{\ensuremath{#1{\scriptstyle\,\pm\,#2}}}
\providecommand{\beststd}[2]{\ensuremath{\mathbf{#1}{\scriptstyle\,\pm\,#2}}}
\providecommand{\secondstd}[2]{\ensuremath{\underline{#1{\scriptstyle\,\pm\,#2}}}}

\begin{table*}[t]
\centering
\scriptsize
\setlength{\tabcolsep}{1.2pt}
\resizebox{\textwidth}{!}{%
\begin{tabular}{lcccccccc}
\toprule
& \multicolumn{4}{c}{Error Correction} & \multicolumn{2}{c}{Step Ordering} & \multicolumn{2}{c}{Protocol Question Answering} \\
\cmidrule(lr){2-5} \cmidrule(lr){6-7} \cmidrule(lr){8-9}
Model & Accuracy & Macro Prec. & Macro Rec. & Macro F1 & Exact Match & Kendall & Accuracy & Brier $\downarrow$ \\
\midrule
\texttt{DeepSeek-V4-Pro}                    & \stdcell{0.67}{0.009} & \stdcell{0.67}{0.009} & \stdcell{0.67}{0.009} & \stdcell{0.67}{0.009} & \stdcell{0.57}{0.010} & \stdcell{0.82}{0.008} & \stdcell{0.70}{0.007} & \stdcell{0.19}{0.003} \\
\texttt{Intern-S1}                          & \stdcell{0.64}{0.007} & \stdcell{0.67}{0.008} & \stdcell{0.64}{0.007} & \stdcell{0.63}{0.007} & \stdcell{0.46}{0.006} & \stdcell{0.72}{0.006} & \stdcell{0.65}{0.009} & \stdcell{0.23}{0.005} \\
\texttt{Intern-S1-Pro}                      & \stdcell{0.65}{0.013} & \stdcell{0.65}{0.013} & \stdcell{0.65}{0.013} & \stdcell{0.65}{0.013} & \stdcell{0.62}{0.004} & \stdcell{0.82}{0.005} & \stdcell{0.70}{0.006} & \stdcell{0.21}{0.003} \\
\texttt{Intern-S2-Preview}                  & \stdcell{0.65}{0.005} & \stdcell{0.67}{0.006} & \stdcell{0.64}{0.005} & \stdcell{0.63}{0.005} & \stdcell{0.52}{0.006} & \stdcell{0.68}{0.016} & \stdcell{0.61}{0.009} & \stdcell{0.26}{0.006} \\
\midrule
\texttt{Qwen2.5-7B}                         & \stdcell{0.56}{0.007} & \stdcell{0.60}{0.012} & \stdcell{0.56}{0.007} & \stdcell{0.50}{0.009} & \stdcell{0.22}{0.014} & \stdcell{0.24}{0.016} & \stdcell{0.46}{0.006} & \stdcell{0.40}{0.004} \\
\texttt{Qwen2.5-7B-LoRA}                    & \stdcell{0.57}{0.008} & \stdcell{0.58}{0.008} & \stdcell{0.57}{0.008} & \stdcell{0.57}{0.008} & \stdcell{0.30}{0.011} & \stdcell{0.45}{0.019} & \stdcell{0.55}{0.012} & \stdcell{0.41}{0.011} \\
\texttt{Qwen2.5-7B-Memory}                  & \stdcell{0.56}{0.009} & \beststd{0.64}{0.020} & \stdcell{0.56}{0.009} & \stdcell{0.49}{0.011} & \stdcell{0.22}{0.010} & \stdcell{0.24}{0.016} & \stdcell{0.53}{0.010} & \beststd{0.36}{0.007} \\
\texttt{Qwen2.5-7B-Memory-LoRA}             & \stdcell{0.59}{0.015} & \stdcell{0.60}{0.015} & \stdcell{0.59}{0.015} & \stdcell{0.59}{0.015} & \stdcell{0.24}{0.014} & \stdcell{0.36}{0.017} & \stdcell{0.55}{0.009} & \stdcell{0.38}{0.008} \\
\texttt{Knowledge Boundary-aware Router}    & \beststd{0.60}{0.014} & \stdcell{0.60}{0.015} & \beststd{0.60}{0.014} & \beststd{0.60}{0.014} & \beststd{0.30}{0.011} & \beststd{0.45}{0.020} & \beststd{0.56}{0.009} & \stdcell{0.38}{0.008} \\
\midrule
\texttt{Qwen3-8B}                           & \stdcell{0.60}{0.006} & \stdcell{0.60}{0.006} & \stdcell{0.60}{0.006} & \stdcell{0.60}{0.011} & \stdcell{0.39}{0.007} & \stdcell{0.63}{0.013} & \stdcell{0.60}{0.006} & \beststd{0.25}{0.004} \\
\texttt{Qwen3-8B-LoRA}                      & \stdcell{0.59}{0.010} & \stdcell{0.59}{0.010} & \stdcell{0.59}{0.010} & \stdcell{0.59}{0.010} & \stdcell{0.39}{0.007} & \stdcell{0.60}{0.015} & \stdcell{0.60}{0.008} & \stdcell{0.32}{0.008} \\
\texttt{Qwen3-8B-Memory}                    & \stdcell{0.61}{0.007} & \stdcell{0.64}{0.009} & \stdcell{0.61}{0.007} & \stdcell{0.58}{0.008} & \beststd{0.43}{0.006} & \beststd{0.69}{0.003} & \stdcell{0.60}{0.005} & \stdcell{0.26}{0.004} \\
\texttt{Qwen3-8B-Memory-LoRA}               & \stdcell{0.62}{0.013} & \stdcell{0.63}{0.013} & \stdcell{0.62}{0.013} & \stdcell{0.61}{0.014} & \stdcell{0.40}{0.009} & \stdcell{0.66}{0.012} & \stdcell{0.39}{0.008} & \stdcell{0.50}{0.006} \\
\texttt{Knowledge Boundary-aware Router}    & \beststd{0.63}{0.007} & \beststd{0.64}{0.007} & \beststd{0.63}{0.007} & \beststd{0.62}{0.008} & \stdcell{0.41}{0.008} & \stdcell{0.66}{0.009} & \beststd{0.61}{0.012} & \stdcell{0.32}{0.010} \\
\bottomrule
\end{tabular}}
\caption{Overall results on BioProBench. Higher is better for all metrics except
Brier score ($\downarrow$). When available, repeated-run variation is reported
as mean $\pm$ standard deviation. Bold indicates the best result within each
Qwen-family block in each column. Lower is better for Brier score.}
\label{tab:bioprobench-overall}
\end{table*}


\subsection{Knowledge-Circuit View of Internal Knowledge Boundaries}
{
Lookup-based conditional memory augments the backbone hidden states by
injecting context-dependent values retrieved from an external store.
Its effect depends on the injection location.
}
For an example $x_i$, let $\mathbf{r}_{i,\tau}$ denote the reasoning state at step $\tau$, represented by hidden state $\mathbf{h}_{i,\tau,l,s}$ at layer $l$ and inference stage $s\in\{\mathrm{pre},\mathrm{dec}\}$. 

We define each layer--stage pair $v=(l,s)$ as a knowledge-circuit node with the local update
\begin{equation}
\begin{array}{rcl}
\mathbf{z}_{i,\tau,l,s} & = & \mathrm{Query}_{l,s}(\mathbf{h}_{i,\tau,l,s}),\\
\mathbf{m}_{i,\tau,l,s} & = & \mathrm{Mem}(\mathbf{z}_{i,\tau,l,s}),\\
\Delta\mathbf{m}_{i,\tau,l,s}
& = & \mathcal{P}_{l,s}(\mathbf{m}_{i,\tau,l,s}),\\
\widehat{\mathbf{h}}_{i,\tau,l,s}
& = & \mathbf{h}_{i,\tau,l,s}+g_{i,\tau,l,s}\Delta\mathbf{m}_{i,\tau,l,s},\\
\mathbf{r}_{i,\tau+1} & = & \mathcal{F}_{l,s}(\widehat{\mathbf{h}}_{i,\tau,l,s}).
\end{array}
\end{equation}

{
Here $\mathrm{Query}_{l,s}$ produces a memory query, $\mathbf{m}_{i,\tau,l,s}$ is the retrieved value, $\mathcal{P}_{l,s}$ projects it to the injected residual $\Delta\mathbf{m}_{i,\tau,l,s}$, and $g_{i,\tau,l,s}$ is the native memory gate. The subsequent backbone computation is denoted by $\mathcal{F}_{l,s}$. Prefill and decoding at the same layer are distinct nodes; thus, $(l,s)$ is the intervention unit selected and scaled by the internal router in Section~4.2.
}

{
This factorization enables controlled causal attenuation of a memory contribution
while leaving the remaining backbone computation fixed. It also separates
availability from usefulness: even a relevant retrieved value may not help at
every node, because its effect depends on the state and stage at which it is
injected. The node-level representation therefore distinguishes a helpful
retrieved association from a harmful or ineffective injection site.
}

\begin{table*}[t]
\centering
\scriptsize
\setlength{\tabcolsep}{1.2pt}
\resizebox{\textwidth}{!}{%
\begin{tabular}{lccccccc}
\toprule
& \multicolumn{1}{c}{Editing} & \multicolumn{3}{c}{Optimization}
& \multicolumn{3}{c}{Understanding} \\
\cmidrule(lr){2-2} \cmidrule(lr){3-5} \cmidrule(lr){6-8}
Model & Accuracy & Mean Imp. & Success Rate & Extraction Rate & MAE $\downarrow$ & Accuracy & TMS \\
\midrule
\texttt{DeepSeek-V4-Pro}     & \stdcell{0.88}{0.098} & \stdcell{0.04}{0.010} & \stdcell{0.08}{0.016} & \stdcell{0.09}{0.015} & \stdcell{0.47}{0.221} & \stdcell{0.62}{0.116} & \stdcell{0.67}{0.577} \\
\texttt{Intern-S1}           & \stdcell{0.84}{0.061} & \stdcell{0.34}{0.031} & \stdcell{0.58}{0.036} & \stdcell{0.79}{0.029} & \stdcell{0.34}{0.028} & \stdcell{0.53}{0.016} & \stdcell{0.33}{0.015} \\
\texttt{Intern-S1-Pro}       & \stdcell{0.90}{0.076} & \stdcell{0.11}{0.021} & \stdcell{0.26}{0.055} & \stdcell{0.41}{0.056} & \stdcell{0.43}{0.142} & \stdcell{0.59}{0.072} & \stdcell{0.39}{0.122} \\
\texttt{Intern-S2-Preview}   & \stdcell{0.85}{0.016} & \stdcell{0.28}{0.023} & \stdcell{0.70}{0.023} & \stdcell{0.96}{0.009} & \stdcell{0.86}{0.029} & \stdcell{0.61}{0.028} & \stdcell{0.70}{0.045} \\
\midrule
\texttt{Qwen2.5-7B}                    & \stdcell{0.27}{0.073} & \stdcell{0.04}{0.043} & \stdcell{0.23}{0.032} & \stdcell{0.89}{0.026} & \stdcell{0.66}{0.063} & \beststd{0.63}{0.036} & \stdcell{0.10}{0.018} \\
\texttt{Qwen2.5-7B-LoRA}               & \stdcell{0.48}{0.110} & \stdcell{0.15}{0.054} & \stdcell{0.34}{0.041} & \stdcell{0.95}{0.020} & \stdcell{0.57}{0.069} & \stdcell{0.53}{0.036} & \beststd{0.27}{0.039} \\
\texttt{Qwen2.5-7B-Memory}             & \stdcell{0.31}{0.064} & \stdcell{0.09}{0.048} & \stdcell{0.31}{0.045} & \beststd{0.98}{0.015} & \stdcell{0.61}{0.081} & \stdcell{0.56}{0.052} & \stdcell{0.14}{0.033} \\
\texttt{Qwen2.5-7B-Memory-LoRA}        & \stdcell{0.46}{0.085} & \stdcell{0.14}{0.031} & \stdcell{0.35}{0.041} & \stdcell{0.95}{0.020} & \stdcell{0.56}{0.129} & \stdcell{0.53}{0.048} & \stdcell{0.24}{0.060} \\
\texttt{Knowledge Boundary-aware Router}  & \beststd{0.51}{0.087} & \beststd{0.16}{0.034} & \beststd{0.36}{0.032} & \stdcell{0.95}{0.019} & \beststd{0.50}{0.052} & \stdcell{0.57}{0.035} & \stdcell{0.26}{0.029} \\
\midrule
\texttt{Qwen3-8B}            & \stdcell{0.43}{0.074} & \stdcell{0.04}{0.029} & \stdcell{0.13}{0.026} & \stdcell{0.84}{0.037} & \stdcell{0.52}{0.071} & \stdcell{0.52}{0.043} & \stdcell{0.16}{0.020} \\
\texttt{Qwen3-8B-LoRA}       & \stdcell{0.56}{0.118} & \stdcell{0.20}{0.028} & \stdcell{0.46}{0.045} & \stdcell{0.98}{0.013} & \stdcell{0.58}{0.064} & \stdcell{0.61}{0.039} & \beststd{0.33}{0.055} \\
\texttt{Qwen3-8B-Memory}     & \stdcell{0.43}{0.111} & \stdcell{0.08}{0.020} & \stdcell{0.19}{0.039} & \stdcell{0.88}{0.028} & \beststd{0.37}{0.072} & \stdcell{0.53}{0.042} & \stdcell{0.24}{0.047} \\
\texttt{Qwen3-8B-Memory-LoRA}& \stdcell{0.50}{0.116} & \stdcell{0.20}{0.037} & \stdcell{0.45}{0.044} & \stdcell{0.98}{0.012} & \stdcell{0.56}{0.122} & \stdcell{0.55}{0.035} & \stdcell{0.29}{0.075} \\
\texttt{Knowledge Boundary-aware Router} & \beststd{0.59}{0.111} & \beststd{0.23}{0.023} & \beststd{0.47}{0.044} & \beststd{0.99}{0.015} & \stdcell{0.48}{0.071} & \beststd{0.69}{0.032} & \stdcell{0.31}{0.049} \\
\bottomrule
\end{tabular}}
\caption{Overall results on ChemCoTBench. Higher is better for all metrics
except MAE ($\downarrow$). When available, repeated-run variation is reported
as mean $\pm$ standard deviation. Bold indicates the best result within each
Qwen-family block in each column. Lower is better for MAE.}
\label{tab:chemcotbench-overall}
\end{table*}

\section{Knowledge Boundary-Aware Routing for Conditional Memory}

{
Section~3 provides two complementary analyses: behavioral boundary codes reveal when the memory-free base model encounters difficulty and layer-stage
knowledge-circuit nodes identify where memory can alter its computation.
The behavioral signals can require reference labels or repeated inference, while the state--node conditions are unavailable before a routed forward pass. We therefore use these analyses to calibrate empirical routing components on existing reasoning samples rather than applying the boundary signals directly at test time.
}

{
In this section, the \textbf{External Boundary-Aware Data Router} converts task-specific, pre-inference input proxies into a global memory-access decision.
Conditional on that decision, the \textbf{Internal Boundary-Aware Parameter Router} configures layer-stage knowledge-circuit nodes and their contribution strengths.
Together, the two components determine whether, where, and how strongly memory participates in a single configured forward path.
}

\subsection{External Boundary-Aware Data Routing for Memory Activation}

For an input $x_i$ from task $t$, we extract a task-specific feature vector organized according to the three boundary types:
\begin{equation}
\vec{\phi}_i^{(t)}=\Phi_t(x_i)
=[\mathbf{p}_{i,A}^{(t)};\mathbf{p}_{i,R}^{(t)};
\mathbf{p}_{i,P}^{(t)}].
\end{equation}
For $h\in\{A,R,P\}$, we write
$\mathbf{p}_{i,h}^{(t)}=[f_{i,h,1}^{(t)},\ldots,f_{i,h,d_{t,h}}^{(t)}]$.
The groups $\mathbf{p}_{i,A}^{(t)}$, $\mathbf{p}_{i,R}^{(t)}$, and
$\mathbf{p}_{i,P}^{(t)}$ encode task-specific proxies for knowledge demand,
reasoning structure, and input ambiguity, respectively.

For $h\in\{A,R,P\}$, we quantize each continuous proxy into task-specific
quantile bins while retaining categorical proxies in their original form:
\begin{equation}
\widetilde f_{i,h,j}^{(t)}=Q_{t,h,j}(f_{i,h,j}^{(t)})
\in\{q_1,\ldots,q_K\}.
\end{equation}
This {knowledge boundary-aware data quantization}  produces the quantized feature vector $\widetilde{\vec{\phi}}_i^{(t)}$.

The external router scores each input using additive bucket contributions and a
small set of explicit feature interactions:
\begin{equation}
\begin{array}{rcl}
\rho_t(x_i) & = & \beta_t^{(0)}
+\displaystyle\sum_{h\in\{A,R,P\}}\sum_{j=1}^{d_{t,h}}
w_{t,h,j}(\widetilde f_{i,h,j}^{(t)})\\
&&+\displaystyle\sum_{r=1}^{R_t}\lambda_{t,r}
\mathbf{1}[C_{t,r}(\widetilde{\vec{\phi}}_i^{(t)})].
\end{array}
\end{equation}
The calibrated external gate is
\begin{equation}
\pi_t(x_i)=
\left\{
\begin{array}{ll}
1, & \rho_t(x_i)\geq\theta_t,\\
0, & \rho_t(x_i)<\theta_t.
\end{array}
\right.
\end{equation}
We tune $\beta_t^{(0)}$, the bucket contributions, interaction rules, and
$\theta_t$ by hyperparameter search using associations between quantized regions
and calibration boundary codes on existing reasoning samples. No additional
classifier is trained. At inference, $\pi_t(x_i)$ uses only input-side proxies;
parameters are selected separately for each task and backbone family.

\subsection{Internal Boundary-Aware Parameter Routing for Memory Activation}

Let $\mathcal{L}_{\mathrm{mem}}$ be the set of memory-enabled layers and let $s\in\{\mathrm{pre},\mathrm{dec}\}$ denote the prefill or decoding stage. Their Cartesian product defines the candidate knowledge-circuit nodes:
\begin{equation}
\mathcal{V}=\mathcal{L}_{\mathrm{mem}}
\times\{\mathrm{pre},\mathrm{dec}\}.
\end{equation}
For an input $x_i$ from task $t$, we form the pre-inference context
\begin{equation}
\mathbf{c}_i^{(t)}=[\mathrm{TaskID}(t);\widetilde{\vec{\phi}}_i^{(t)}].
\end{equation}
The internal router maps this context to
\begin{equation}
\vec{\eta}_i^{(t)}=R_{\mathrm{par}}(\mathbf{c}_i^{(t)})
=\left(\mathbf{u}_i^{L},\mathbf{u}_i^{S},
\mathbf{a}_i\right),
\end{equation}
where $\mathbf{u}_i^{L}\in\{0,1\}^{|\mathcal{L}_{\mathrm{mem}}|}$ selects
layers, $\mathbf{u}_i^{S}\in\{0,1\}^{2}$ selects stages, and
$\mathbf{a}_i=\{a_{i,l,s}\}_{(l,s)\in\mathcal{V}}$ with
$a_{i,l,s}\in[0,1]$ scales individual nodes. Thus, a zero mask or coefficient
suppresses its corresponding memory contribution.

At node $(l,s)$ and reasoning step $\tau$, the routed hidden-state update is
\begin{equation}
\begin{array}{rcl}
\widehat{\mathbf{h}}_{i,\tau,l,s} & = & \mathbf{h}_{i,\tau,l,s}
+\pi_t(x_i)\,u_{i,l}^{L}u_{i,s}^{S}\\
&&\cdot a_{i,l,s}g_{i,\tau,l,s}
\Delta\mathbf{m}_{i,\tau,l,s},
\end{array}
\end{equation}
where $\Delta\mathbf{m}_{i,\tau,l,s}$ and $g_{i,\tau,l,s}$ are the native
memory residual and gate, respectively. We derive $R_{\mathrm{par}}$ through
causal attenuation interventions at candidate nodes on existing reasoning samples
and select its parameters by hyperparameter search. At inference, the router
instantiates $\vec{\eta}_i^{(t)}$ from $\mathbf{c}_i^{(t)}$ without reference
answers or alternative model outputs, while leaving backbone and memory parameters
unchanged.

\section{Experimental Setup}
\subsection{Datasets}
{
We evaluate conditional memory on two  scientific reasoning benchmarks. \textbf{BioProBench}~\cite{DBLP:journals/corr/abs-2505-07889} is grounded in human-authored biological protocols and evaluates procedural reasoning. 
We use Protocol Question Answering (PQA), Step Ordering (ORD), and Error Correction (ERR) to assess retrieval of procedural facts, reconstruction of causal step dependencies, and identification of safety- or validity-critical errors. 

\textbf{ChemCoTBench}~\cite{DBLP:journals/corr/abs-2505-21318} evaluates step-wise chemical reasoning through explicit, verifiable molecular operations. 
We focus on molecular understanding (MolUnd), molecule editing (MolEdit), and molecular optimization (MolOpt), covering structure comprehension, instruction-guided modification, and property-guided design.
We use the official test splits and task-specific evaluators.
}

\subsection{Models and Baselines}
{
We use Qwen2.5-7B and Qwen3-8B as routable backbone families. For each family and benchmark, we compare the base model, its domain-adapted LoRA variant, a memory-only variant, and Memory+LoRA. Guided by the knowledge-boundary characterization in Section~3.1 and the knowledge-circuit view operationalized in Section~4.2, we construct two complementary empirical routing mechanisms from existing reasoning samples. Their task- and backbone-specific parameters are obtained through hyperparameter search and applied at inference to configure memory access and knowledge-circuit contributions for each input. The \textbf{External Boundary-Aware Data Router} derives an empirical memory-access configuration from pre-generation, task-specific input features, while the \textbf{Internal Boundary-Aware Parameter Router} configures active knowledge-circuit nodes and their intervention strengths. Together, they specify when, where, and how strongly memory participates in computation. We additionally report strong non-routable API models as reference systems. No score or threshold is transferred from Qwen2.5 to Qwen3.
}

\subsection{Evaluation Metrics}
{
To characterize knowledge boundaries, we measure answer failure (AF), reasoning-step failure (RF), and prediction instability (PI) with respect to the memory-off base model. These three signals define the analysis cohorts for the boundary and ablation studies.
}

{
For task-level performance, BioProBench ERR uses accuracy and macro precision, recall, and F1. ORD uses exact match and Kendall's $\tau$. PQA uses accuracy and Brier score. For ChemCoTBench, MolEdit uses editing accuracy. MolOpt uses mean property improvement, success rate, and extraction rate. MolUnd uses mean absolute error (MAE), accuracy, and Tanimoto molecular similarity (TMS). Higher is better except for Brier score and MAE. We follow the official evaluators and report parsing and validity details in the supplementary material.
}


\section{Main Results}
\subsection{BioProBench: Procedural Reasoning Results}
{
Table~\ref{tab:bioprobench-overall} shows that unconditional memory is not uniformly beneficial for biological procedural reasoning. In the Qwen2.5 block, memory alone attains the best macro precision and Brier score, but does not improve ordering; the Knowledge Boundary-aware Router instead yields the highest ERR accuracy, recall, and F1, matches the LoRA ordering scores, and achieves the highest PQA accuracy. For Qwen3, Memory+LoRA increases ERR accuracy but sharply degrades PQA (0.39 accuracy and 0.50 Brier), whereas the router achieves the strongest ERR and PQA accuracies within the family (0.63 and 0.61) while retaining LoRA-level calibration. Thus, the gains arise from conditioning memory on the input rather than treating it as a uniformly useful augmentation.
}

\subsection{ChemCoTBench: Chemical Reasoning Results}
{
Table~\ref{tab:chemcotbench-overall} reveals the same conditional pattern for chemical reasoning. In the Qwen2.5 block, the router obtains the best editing accuracy (0.51), mean property improvement (0.16), success rate (0.36), and MAE (0.50), while memory alone has the highest extraction rate and non-routed baselines retain the best molecular-understanding accuracy and TMS. In the Qwen3 block, the router leads editing, all three optimization metrics, and understanding accuracy, but the memory-only model retains the lowest MAE and LoRA the highest TMS. No fixed memory configuration therefore dominates the full metric suite: memory should be directed to inputs and knowledge-circuit configurations where it improves the task-specific scientific objective, which is precisely what the Knowledge Boundary-aware Router targets.
}

\FloatBarrier
\section{Ablation Study}

\subsection{Boundary-Dependent Memory Effects}
\begin{figure}[t]
    \centering
    \includegraphics[width=\linewidth]{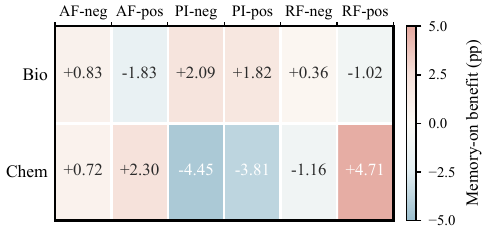}
    \caption{Knowledge-boundary ablation. Percentage-point effects of enabling memory across AF, PI, and RF conditions in BioProBench and ChemCoTBench. Positive values indicate gains and negative values regressions; AF, PI, and RF denote final-answer failure, prediction instability, and reasoning failure.}
    \label{fig:knowledge-boundary-ablation}
\end{figure}

{
Figure~\ref{fig:knowledge-boundary-ablation} establishes the premise of conditional memory. The effect of enabling memory changes sign across boundary cohorts and benchmarks: Bio prediction-instability cohorts improve by 2.09 and 1.82 points, whereas the corresponding Chem cohorts decline by 4.45 and 3.81 points; Chem reasoning-failure-positive examples, in contrast, improve by 4.71 points. These mixed effects identify both regimes where memory repairs missing scientific knowledge and regimes where it introduces interference, motivating conditional rather than uniform memory use.
}

\begin{table}[t]
\centering
\scriptsize
\setlength{\tabcolsep}{2.0pt}
\resizebox{\linewidth}{!}{%
\begin{tabular}{lcccc}
\toprule
Model & Accuracy & Macro-P & Macro-R & Macro-F1 \\
\midrule
Base & \stdcell{0.556}{0.007} & \stdcell{0.602}{0.012} & \stdcell{0.555}{0.007} & \beststd{0.498}{0.009} \\
Engram-only & \beststd{0.562}{0.009} & \beststd{0.637}{0.020} & \beststd{0.561}{0.009} & \stdcell{0.490}{0.010} \\
MeKi-only & \stdcell{0.531}{0.006} & \stdcell{0.573}{0.052} & \stdcell{0.506}{0.004} & \stdcell{0.372}{0.004} \\
\bottomrule
\end{tabular}}
\caption{ERR diagnostic for Base, Engram-only, and MeKi-only on
BioProBench. Values are means $\pm$ standard deviations; higher is better.
All results aggregate three repeated runs.}
\label{tab:meki-engram-err}
\end{table}

\subsection{ {Memory-Adapter Architecture Comparison}}
{
To contextualize the effects of the Engram adapter~\cite{DBLP:conf/acl/ChengZDCWXHYHZL26}, we compare it with the memory-expert architecture MeKi~\cite{ding2026meki} on BioProBench Error Correction (ERR). 
Table~\ref{tab:meki-engram-err} reports the results over three repeated runs. 
Overall, Engram attains a higher parsing success rate than MeKi. 
Together with the ERR outcomes, this observation motivates examining how their retrieval and residual-injection paths differ. 
The supplementary material provides a structural comparison of the two adapters.
}

\subsection{External Routing versus Activation-Rate-Matched Random Routing}
\begin{figure*}[!t]
    \centering

    \includegraphics[width=0.92\textwidth]
    {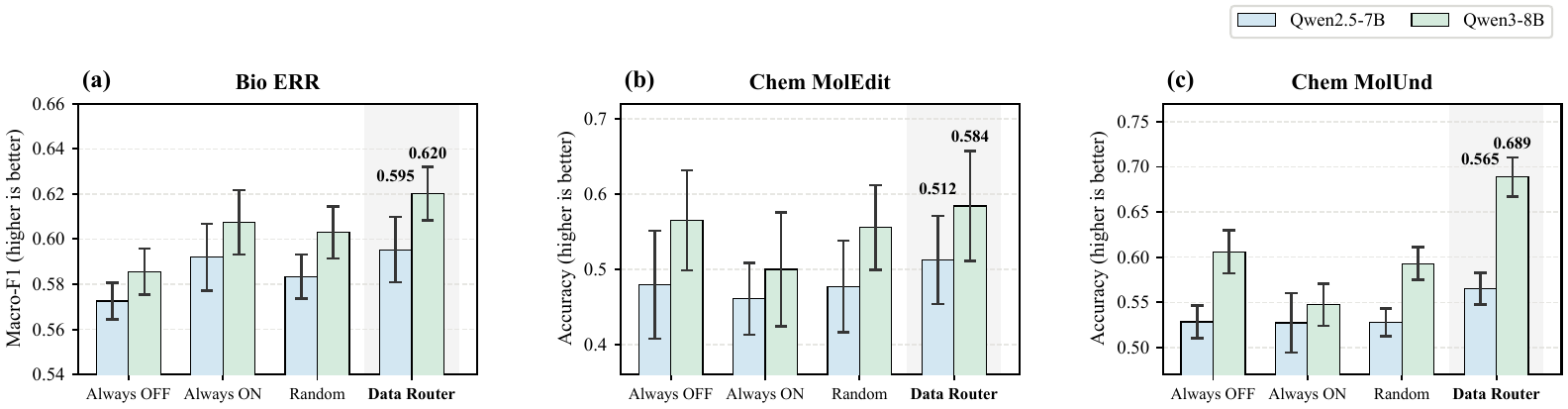}
    \caption{External data-router ablation on Bio ERR and two ChemCoTBench tasks.
    We compare Always OFF, Always ON, a Random Router
    matched to the Data Router's memory-on rate, and the empirical
    feature-based Data Router for Qwen2.5-7B and Qwen3-8B. Bars and
    error bars show the mean and standard deviation over repeated runs;
    Random Router results additionally average over 100 fixed masks.
    The Data Router outperforms the matched random baseline in every
    setting shown, indicating that the gains depend on which samples
    activate memory rather than the activation rate alone.}
    \label{fig:external-data-router-ablation}

    \includegraphics[width=0.92\textwidth]
    {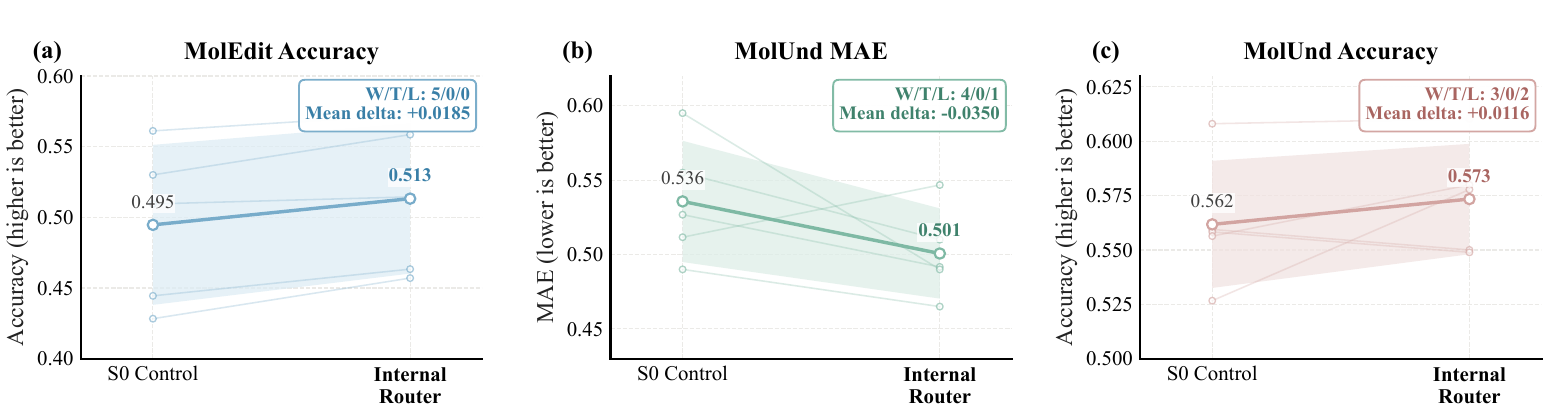}
    \caption{Internal memory-router ablation on Qwen2.5-7B. On the fixed
    screen cohort selected by the external Data Router, we compare the
    empirical task-conditioned router with an S0 control using the same
    cohort, runs, and generation batch. Thin lines show paired
    runs, thick lines show their means, and shaded regions denote one
    standard deviation; W/T/L counts run-wise wins, ties, and losses.
    The empirical router improves MolEdit accuracy and MolUnd MAE and
    accuracy. Because this is a same-cohort offline replay, the result
    validates the routing mechanism rather than independent
    generalization.}
    \label{fig:internal-router-ablation}

    \includegraphics[width=0.92\textwidth]
    {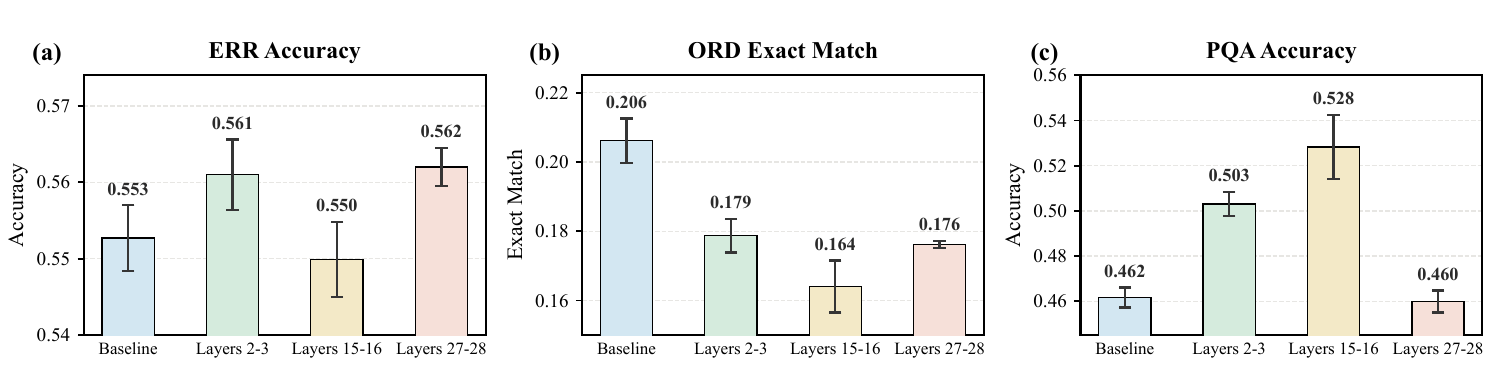}
    \caption{Memory-layer ablation on BioProBench. Bars report task
    metrics conditioned on successful parsing, and error bars denote
    standard deviations over repeated runs. Layer numbers are one-based.}
    \label{fig:memory-layer-ablation}
\end{figure*}

{
Figure~\ref{fig:external-data-router-ablation} isolates whether the external router benefits merely from its overall memory-access rate. To cover both benchmarks, we display Bio ERR from BioProBench together with the two ChemCoTBench settings with the largest average Data Router gain over the activation-rate-matched Random Router across both backbone families: MolEdit and MolUnd. The empirical feature-based Data Router outperforms the Random Router in every displayed setting and provides a stronger trade-off than the fixed Always OFF and Always ON configurations. These comparisons show that the gains depend on assigning memory access to appropriate inputs, rather than on activating memory more frequently, and motivate task- and backbone-specific routing rules.
}

\subsection{Internal Routing of Knowledge-Circuit Nodes}
{
Figure~\ref{fig:internal-router-ablation} evaluates the Internal Boundary-Aware Parameter Router on the fixed cohort selected by the external router, using the same runs and generation batches as the S0 control. For Qwen2.5-7B, the empirical internal configuration improves MolEdit accuracy from 0.495 to 0.513 and reduces MolUnd MAE from 0.536 to 0.501, while also increasing MolUnd accuracy from 0.562 to 0.573. The effect is metric-dependent rather than uniformly positive, which is consistent with the knowledge-circuit view: configuring memory at selected layer--stage knowledge-circuit nodes can preserve useful contributions while attenuating harmful ones. Because this is a same-cohort offline replay, it provides mechanism-level validation rather than an independent generalization estimate.
}

\subsection{Task-Dependent Memory Placement}
{
Figure~\ref{fig:memory-layer-ablation} shows that a globally fixed injection location is also inadequate. Late injection at actual layers 27--28 gives the strongest ERR accuracy (0.562), middle injection at layers 15--16 gives the highest PQA accuracy (0.528), and the memory-free baseline remains best for ORD exact match (0.206). The PQA scores are conditioned on successful parsing and should therefore be read together with the lower parsing coverage of the middle-layer configuration. These task-specific optima support the internal router's layer--stage control: both whether memory participates and where it enters the computation should depend on the downstream reasoning objective.
}

\section{Conclusion}
{
Memory is not always needed for scientific reasoning: its value depends on whether the base model reaches a knowledge boundary and whether retrieved signals enter at knowledge-circuit nodes that can use them. 
We characterize this dependence through behavioral indicators of answer, reasoning, and stability failures together with controlled layer-stage interventions. 
These analyses motivate a Knowledge Boundary-Aware Router that uses pre-inference input proxies to control global memory access and internal node contributions without reference answers or alternative outputs. 
Across two benchmarks, two backbone families, and six task types, memory effects vary across inputs, tasks, metrics, and injection locations. 
Feature-based routing beats activation-rate-matched random routing, while targeted internal
routing attenuates harmful contributions. 
Conditional memory should therefore be treated as a selective computational resource rather than a uniformly beneficial augmentation.
}

\bibliography{aaai2027}

@article{DBLP:journals/natmi/JablonkaSOS24,
  author    = {Kevin Maik Jablonka and
               Philippe Schwaller and
               Andres Ortega{-}Guerrero and
               Berend Smit},
  title     = {Leveraging large language models for predictive chemistry},
  journal   = {Nat. Mac. Intell.},
  volume    = {6},
  number    = {2},
  pages     = {161--169},
  year      = {2024},
  url       = {https://doi.org/10.1038/s42256-023-00788-1},
  doi       = {10.1038/S42256-023-00788-1},
  bibsource = {dblp computer science bibliography, https://dblp.org}
}

@article{DBLP:journals/natmi/RossBCPMD22,
  author    = {Jerret Ross and
               Brian Belgodere and
               Vijil Chenthamarakshan and
               Inkit Padhi and
               Youssef Mroueh and
               Payel Das},
  title     = {Large-scale chemical language representations capture molecular structure
               and properties},
  journal   = {Nat. Mac. Intell.},
  volume    = {4},
  number    = {12},
  pages     = {1256--1264},
  year      = {2022},
  url       = {https://doi.org/10.1038/s42256-022-00580-7},
  doi       = {10.1038/S42256-022-00580-7},
  bibsource = {dblp computer science bibliography, https://dblp.org}
}

@article{DBLP:journals/nature/MerchantBSACC23,
  author    = {Amil Merchant and
               Simon L. Batzner and
               Samuel S. Schoenholz and
               Muratahan Aykol and
               Gowoon Cheon and
               Ekin Dogus Cubuk},
  title     = {Scaling deep learning for materials discovery},
  journal   = {Nat.},
  volume    = {624},
  number    = {7990},
  pages     = {80--85},
  year      = {2023},
  url       = {https://doi.org/10.1038/s41586-023-06735-9},
  doi       = {10.1038/S41586-023-06735-9},
  bibsource = {dblp computer science bibliography, https://dblp.org}
}

@article{DBLP:journals/nature/BoikoMKG23,
  author    = {Daniil A. Boiko and
               Robert MacKnight and
               Ben Kline and
               Gabe Gomes},
  title     = {Autonomous chemical research with large language models},
  journal   = {Nat.},
  volume    = {624},
  number    = {7992},
  pages     = {570--578},
  year      = {2023},
  url       = {https://doi.org/10.1038/s41586-023-06792-0},
  doi       = {10.1038/S41586-023-06792-0},
  bibsource = {dblp computer science bibliography, https://dblp.org}
}

@article{DBLP:journals/nature/RomeraParedesBNBKDREWFKF24,
  author    = {Bernardino Romera{-}Paredes and
               Mohammadamin Barekatain and
               Alexander Novikov and
               Matej Balog and
               M. Pawan Kumar and
               Emilien Dupont and
               Francisco J. R. Ruiz and
               Jordan S. Ellenberg and
               Pengming Wang and
               Omar Fawzi and
               Pushmeet Kohli and
               Alhussein Fawzi},
  title     = {Mathematical discoveries from program search with large language models},
  journal   = {Nat.},
  volume    = {625},
  number    = {7995},
  pages     = {468--475},
  year      = {2024},
  url       = {https://doi.org/10.1038/s41586-023-06924-6},
  doi       = {10.1038/S41586-023-06924-6},
  bibsource = {dblp computer science bibliography, https://dblp.org}
}

@inproceedings{DBLP:conf/nips/ZhangHZDYWYD024,
  author    = {Dan Zhang and
               Ziniu Hu and
               Sining Zhoubian and
               Zhengxiao Du and
               Kaiyu Yang and
               Zihan Wang and
               Yisong Yue and
               Yuxiao Dong and
               Jie Tang},
  title     = {SciInstruct: a Self-Reflective Instruction Annotated Dataset for Training
               Scientific Language Models},
  booktitle = {Advances in Neural Information Processing Systems 37: Annual Conference
               on Neural Information Processing Systems 2024, NeurIPS 2024, Vancouver,
               BC, Canada, December 10--15, 2024},
  year      = {2024},
  url       = {https://proceedings.neurips.cc/paper_files/paper/2024/hash/02ee6b7295f720407b56c457b34c54d5-Abstract-Datasets_and_Benchmarks_Track.html},
  bibsource = {dblp computer science bibliography, https://dblp.org}
}

@inproceedings{DBLP:conf/iclr/WenTDD0XT24,
  author    = {Hongzhi Wen and
               Wenzhuo Tang and
               Xinnan Dai and
               Jiayuan Ding and
               Wei Jin and
               Yuying Xie and
               Jiliang Tang},
  title     = {CellPLM: Pre-training of Cell Language Model Beyond Single Cells},
  booktitle = {The Twelfth International Conference on Learning Representations,
               {ICLR} 2024, Vienna, Austria, May 7--11, 2024},
  publisher = {OpenReview.net},
  year      = {2024},
  url       = {https://openreview.net/forum?id=BKXvPDekud},
  bibsource = {dblp computer science bibliography, https://dblp.org}
}

@inproceedings{DBLP:conf/icml/GuuLTPC20,
  author    = {Kelvin Guu and
               Kenton Lee and
               Zora Tung and
               Panupong Pasupat and
               Ming{-}Wei Chang},
  title     = {Retrieval Augmented Language Model Pre-Training},
  booktitle = {Proceedings of the 37th International Conference on Machine Learning,
               {ICML} 2020, 13--18 July 2020, Virtual Event},
  series    = {Proceedings of Machine Learning Research},
  volume    = {119},
  pages     = {3929--3938},
  publisher = {{PMLR}},
  year      = {2020},
  url       = {https://proceedings.mlr.press/v119/guu20a.html},
  bibsource = {dblp computer science bibliography, https://dblp.org}
}

@inproceedings{DBLP:conf/acl/ChengZDCWXHYHZL26,
  author    = {Xin Cheng and
               Wangding Zeng and
               Damai Dai and
               Qinyu Chen and
               Bingxuan Wang and
               Zhenda Xie and
               Kezhao Huang and
               Xingkai Yu and
               Zhewen Hao and
               Han Zhang and
               Yukun Li and
               Huishuai Zhang and
               Dongyan Zhao and
               Wenfeng Liang},
  title     = {Conditional Memory via Scalable Lookup: {A} New Axis of Sparsity for
               Large Language Models},
  booktitle = {Proceedings of the 64th Annual Meeting of the Association for Computational
               Linguistics (Volume 1: Long Papers), {ACL} 2026, San Diego, California,
               United States, July 2-7, 2026},
  pages     = {4968--4990},
  publisher = {Association for Computational Linguistics},
  year      = {2026},
  url       = {https://aclanthology.org/2026.acl-long.226/},
  bibsource = {dblp computer science bibliography, https://dblp.org}
}

@article{DBLP:journals/corr/abs-2601-22203,
  author     = {Huinan Xu and
                Xuyang Feng and
                Junhong Chen and
                Junchen Liu and
                Kaiwen Deng and
                Kai Ding and
                Shengning Long and
                Jiaxue Shuai and
                Zhaorong Li and
                Shiping Liu and
                Guirong Xue and
                Zhan Xiao},
  title      = {Beyond Conditional Computation: Retrieval-Augmented Genomic
                Foundation Models with Gengram},
  journal    = {CoRR},
  volume     = {abs/2601.22203},
  year       = {2026},
  url        = {https://doi.org/10.48550/arXiv.2601.22203},
  doi        = {10.48550/ARXIV.2601.22203},
  eprinttype = {arXiv},
  eprint     = {2601.22203},
  bibsource  = {dblp computer science bibliography, https://dblp.org}
}

@article{DBLP:journals/corr/abs-2606-12113,
  author     = {Xinni Zhang and
                Zijing Liu and
                He Cao and
                Yu Li and
                Irwin King},
  title      = {Augmenting Molecular Language Models with Local {$n$}-gram Memory},
  journal    = {CoRR},
  volume     = {abs/2606.12113},
  year       = {2026},
  url        = {https://doi.org/10.48550/arXiv.2606.12113},
  doi        = {10.48550/ARXIV.2606.12113},
  eprinttype = {arXiv},
  eprint     = {2606.12113},
  bibsource  = {dblp computer science bibliography, https://dblp.org}
}

@article{DBLP:journals/corr/abs-2508-15763,
  author     = {Intern{-}S1 Team and
                Runyu Peng and
                Yu Qiao},
  title      = {Intern-S1: {A} Scientific Multimodal Foundation Model},
  journal    = {CoRR},
  volume     = {abs/2508.15763},
  year       = {2025},
  url        = {https://doi.org/10.48550/arXiv.2508.15763},
  doi        = {10.48550/ARXIV.2508.15763},
  eprinttype = {arXiv},
  eprint     = {2508.15763},
  bibsource  = {dblp computer science bibliography, https://dblp.org}
}

@article{DBLP:journals/corr/abs-2505-07889,
  author     = {Yuyang Liu and
                Liuzhenghao Lv and
                Xiancheng Zhang and
                Li Yuan and
                Yonghong Tian},
  title      = {{BioProBench}: Comprehensive Dataset and Benchmark in Biological
                Protocol Understanding and Reasoning},
  journal    = {CoRR},
  volume     = {abs/2505.07889},
  year       = {2025},
  url        = {https://doi.org/10.48550/arXiv.2505.07889},
  doi        = {10.48550/ARXIV.2505.07889},
  eprinttype = {arXiv},
  eprint     = {2505.07889},
  bibsource  = {dblp computer science bibliography, https://dblp.org}
}

@article{DBLP:journals/corr/abs-2505-21318,
  author     = {Hao Li and
                He Cao and
                Bin Feng and
                Yanjun Shao and
                Xiangru Tang and
                Zhiyuan Yan and
                Li Yuan and
                Yonghong Tian and
                Yu Li},
  title      = {Beyond Chemical {QA}: Evaluating {LLM}'s Chemical Reasoning with
                Modular Chemical Operations},
  journal    = {CoRR},
  volume     = {abs/2505.21318},
  year       = {2025},
  url        = {https://doi.org/10.48550/arXiv.2505.21318},
  doi        = {10.48550/ARXIV.2505.21318},
  eprinttype = {arXiv},
  eprint     = {2505.21318},
  bibsource  = {dblp computer science bibliography, https://dblp.org}
}

@article{ding2026meki,
  author     = {Ning Ding and
                Fangcheng Liu and
                Kyungrae Kim and
                Linji Hao and
                Kyeng{-}Hun Lee and
                Hyeonmok Ko and
                Yehui Tang},
  title      = {{MeKi}: Memory-based Expert Knowledge Injection for Efficient
                {LLM} Scaling},
  journal    = {CoRR},
  volume     = {abs/2602.03359},
  year       = {2026},
  url        = {https://doi.org/10.48550/arXiv.2602.03359},
  doi        = {10.48550/ARXIV.2602.03359},
  eprinttype = {arXiv},
  eprint     = {2602.03359}
}

@inproceedings{DBLP:conf/nips/LampleSRDJ19,
  author    = {Guillaume Lample and
               Alexandre Sablayrolles and
               Marc'Aurelio Ranzato and
               Ludovic Denoyer and
               Herv{\'{e}} J{\'{e}}gou},
  editor    = {Hanna M. Wallach and
               Hugo Larochelle and
               Alina Beygelzimer and
               Florence d'Alch{\'{e}}{-}Buc and
               Emily B. Fox and
               Roman Garnett},
  title     = {Large Memory Layers with Product Keys},
  booktitle = {Advances in Neural Information Processing Systems 32: Annual Conference
               on Neural Information Processing Systems 2019, NeurIPS 2019, December
               8-14, 2019, Vancouver, BC, Canada},
  pages     = {8546--8557},
  year      = {2019},
  url       = {https://proceedings.neurips.cc/paper/2019/hash/9d8df73a3cfbf3c5b47bc9b50f214aff-Abstract.html},
  bibsource = {dblp computer science bibliography, https://dblp.org}
}

@inproceedings{DBLP:conf/iclr/KhandelwalLJZL20,
  author    = {Urvashi Khandelwal and
               Omer Levy and
               Dan Jurafsky and
               Luke Zettlemoyer and
               Mike Lewis},
  title     = {Generalization through Memorization: Nearest Neighbor Language Models},
  booktitle = {8th International Conference on Learning Representations, {ICLR} 2020,
               Addis Ababa, Ethiopia, April 26-30, 2020},
  publisher = {OpenReview.net},
  year      = {2020},
  url       = {https://openreview.net/forum?id=HklBjCEKvH},
  bibsource = {dblp computer science bibliography, https://dblp.org}
}

@inproceedings{DBLP:conf/nips/LewisPPPKGKLYR020,
  author    = {Patrick Lewis and
               Ethan Perez and
               Aleksandra Piktus and
               Fabio Petroni and
               Vladimir Karpukhin and
               Naman Goyal and
               Heinrich K{\"{u}}ttler and
               Mike Lewis and
               Wen{-}tau Yih and
               Tim Rockt{\"{a}}schel and
               Sebastian Riedel and
               Douwe Kiela},
  editor    = {Hugo Larochelle and
               Marc'Aurelio Ranzato and
               Raia Hadsell and
               Maria{-}Florina Balcan and
               Hsuan{-}Tien Lin},
  title     = {Retrieval-Augmented Generation for Knowledge-Intensive {NLP} Tasks},
  booktitle = {Advances in Neural Information Processing Systems 33: Annual Conference
               on Neural Information Processing Systems 2020, NeurIPS 2020, December
               6-12, 2020, virtual},
  year      = {2020},
  url       = {https://proceedings.neurips.cc/paper/2020/hash/6b493230205f780e1bc26945df7481e5-Abstract.html},
  bibsource = {dblp computer science bibliography, https://dblp.org}
}

@inproceedings{DBLP:conf/icml/BorgeaudMHCRM0L22,
  author    = {Sebastian Borgeaud and
               Arthur Mensch and
               Jordan Hoffmann and
               Trevor Cai and
               Eliza Rutherford and
               Katie Millican and
               George van den Driessche and
               Jean{-}Baptiste Lespiau and
               Bogdan Damoc and
               Aidan Clark and
               Diego de Las Casas and
               Aurelia Guy and
               Jacob Menick and
               Roman Ring and
               Tom Hennigan and
               Saffron Huang and
               Loren Maggiore and
               Chris Jones and
               Albin Cassirer and
               Andy Brock and
               Michela Paganini and
               Geoffrey Irving and
               Oriol Vinyals and
               Simon Osindero and
               Karen Simonyan and
               Jack W. Rae and
               Erich Elsen and
               Laurent Sifre},
  editor    = {Kamalika Chaudhuri and
               Stefanie Jegelka and
               Le Song and
               Csaba Szepesv{\'{a}}ri and
               Gang Niu and
               Sivan Sabato},
  title     = {Improving Language Models by Retrieving from Trillions of Tokens},
  booktitle = {International Conference on Machine Learning, {ICML} 2022, 17-23 July
               2022, Baltimore, Maryland, {USA}},
  series    = {Proceedings of Machine Learning Research},
  volume    = {162},
  pages     = {2206--2240},
  publisher = {{PMLR}},
  year      = {2022},
  url       = {https://proceedings.mlr.press/v162/borgeaud22a.html},
  bibsource = {dblp computer science bibliography, https://dblp.org}
}

@inproceedings{DBLP:conf/iclr/WuRHS22,
  author    = {Yuhuai Wu and
               Markus Norman Rabe and
               DeLesley Hutchins and
               Christian Szegedy},
  title     = {Memorizing Transformers},
  booktitle = {The Tenth International Conference on Learning Representations, {ICLR}
               2022, Virtual Event, April 25-29, 2022},
  publisher = {OpenReview.net},
  year      = {2022},
  url       = {https://openreview.net/forum?id=TrjbxzRcnf-},
  bibsource = {dblp computer science bibliography, https://dblp.org}
}

@inproceedings{DBLP:conf/nips/Wang0CLYGW23,
  author    = {Weizhi Wang and
               Li Dong and
               Hao Cheng and
               Xiaodong Liu and
               Xifeng Yan and
               Jianfeng Gao and
               Furu Wei},
  editor    = {Alice Oh and
               Tristan Naumann and
               Amir Globerson and
               Kate Saenko and
               Moritz Hardt and
               Sergey Levine},
  title     = {Augmenting Language Models with Long-Term Memory},
  booktitle = {Advances in Neural Information Processing Systems 36: Annual Conference
               on Neural Information Processing Systems 2023, NeurIPS 2023, New Orleans,
               LA, USA, December 10 - 16, 2023},
  year      = {2023},
  url       = {http://papers.nips.cc/paper\_files/paper/2023/hash/ebd82705f44793b6f9ade5a669d0f0bf-Abstract-Conference.html},
  bibsource = {dblp computer science bibliography, https://dblp.org}
}

@article{DBLP:journals/corr/abs-2312-17257,
  author     = {Ruifeng Yuan and
                Shichao Sun and
                Zili Wang and
                Ziqiang Cao and
                Wenjie Li},
  title      = {Evolving Large Language Model Assistant with Long-Term Conditional
                Memory},
  journal    = {CoRR},
  volume     = {abs/2312.17257},
  year       = {2023},
  url        = {https://doi.org/10.48550/arXiv.2312.17257},
  doi        = {10.48550/ARXIV.2312.17257},
  eprinttype = {arXiv},
  eprint     = {2312.17257},
  bibsource  = {dblp computer science bibliography, https://dblp.org}
}

@inproceedings{DBLP:conf/icml/WangGCJLYYLLYSM24,
  author    = {Yu Wang and
               Yifan Gao and
               Xiusi Chen and
               Haoming Jiang and
               Shiyang Li and
               Jingfeng Yang and
               Qingyu Yin and
               Zheng Li and
               Xian Li and
               Bing Yin and
               Jingbo Shang and
               Julian J. McAuley},
  editor    = {Ruslan Salakhutdinov and
               Zico Kolter and
               Katherine A. Heller and
               Adrian Weller and
               Nuria Oliver and
               Jonathan Scarlett and
               Felix Berkenkamp},
  title     = {{MEMORYLLM:} Towards Self-Updatable Large Language Models},
  booktitle = {Forty-first International Conference on Machine Learning, {ICML} 2024,
               Vienna, Austria, July 21-27, 2024},
  series    = {Proceedings of Machine Learning Research},
  volume    = {235},
  pages     = {50453--50466},
  publisher = {{PMLR} / OpenReview.net},
  year      = {2024},
  url       = {https://proceedings.mlr.press/v235/wang24s.html},
  bibsource = {dblp computer science bibliography, https://dblp.org}
}

@article{DBLP:journals/corr/abs-2403-11901,
  author     = {Payel Das and
                Subhajit Chaudhury and
                Elliot Nelson and
                Igor Melnyk and
                Sarath Swaminathan and
                Sihui Dai and
                Aur{\'{e}}lie C. Lozano and
                Georgios Kollias and
                Vijil Chenthamarakshan and
                Jir{\'{\i}} Navr{\'{a}}til and
                Soham Dan and
                Pin{-}Yu Chen},
  title      = {Larimar: Large Language Models with Episodic Memory Control},
  journal    = {CoRR},
  volume     = {abs/2403.11901},
  year       = {2024},
  url        = {https://doi.org/10.48550/arXiv.2403.11901},
  doi        = {10.48550/ARXIV.2403.11901},
  eprinttype = {arXiv},
  eprint     = {2403.11901},
  bibsource  = {dblp computer science bibliography, https://dblp.org}
}

@inproceedings{DBLP:conf/icml/BergesOHYZG25,
  author    = {Vincent{-}Pierre Berges and
               Barlas Oguz and
               Daniel Haziza and
               Wen{-}tau Yih and
               Luke Zettlemoyer and
               Gargi Ghosh},
  editor    = {Aarti Singh and
               Maryam Fazel and
               Daniel Hsu and
               Simon Lacoste{-}Julien and
               Felix Berkenkamp and
               Tegan Maharaj and
               Kiri Wagstaff and
               Jerry Zhu},
  title     = {Memory Layers at Scale},
  booktitle = {Forty-second International Conference on Machine Learning, {ICML}
               2025, Vancouver, BC, Canada, July 13-19, 2025},
  series    = {Proceedings of Machine Learning Research},
  volume    = {267},
  publisher = {{PMLR} / OpenReview.net},
  year      = {2025},
  url       = {https://proceedings.mlr.press/v267/berges25a.html},
  bibsource = {dblp computer science bibliography, https://dblp.org}
}

@inproceedings{DBLP:conf/iclr/HendrycksBBZMSS21,
  author    = {Dan Hendrycks and
               Collin Burns and
               Steven Basart and
               Andy Zou and
               Mantas Mazeika and
               Dawn Song and
               Jacob Steinhardt},
  title     = {Measuring Massive Multitask Language Understanding},
  booktitle = {9th International Conference on Learning Representations, {ICLR} 2021,
               Virtual Event, Austria, May 3-7, 2021},
  publisher = {OpenReview.net},
  year      = {2021},
  url       = {https://openreview.net/forum?id=d7KBjmI3GmQ},
  bibsource = {dblp computer science bibliography, https://dblp.org}
}

@inproceedings{DBLP:conf/icml/WangHL0ZSLZS024,
  author    = {Xiaoxuan Wang and
               Ziniu Hu and
               Pan Lu and
               Yanqiao Zhu and
               Jieyu Zhang and
               Satyen Subramaniam and
               Arjun R. Loomba and
               Shichang Zhang and
               Yizhou Sun and
               Wei Wang},
  editor    = {Ruslan Salakhutdinov and
               Zico Kolter and
               Katherine A. Heller and
               Adrian Weller and
               Nuria Oliver and
               Jonathan Scarlett and
               Felix Berkenkamp},
  title     = {SciBench: Evaluating College-Level Scientific Problem-Solving Abilities
               of Large Language Models},
  booktitle = {Forty-first International Conference on Machine Learning, {ICML} 2024,
               Vienna, Austria, July 21-27, 2024},
  series    = {Proceedings of Machine Learning Research},
  volume    = {235},
  pages     = {50622--50649},
  publisher = {{PMLR} / OpenReview.net},
  year      = {2024},
  url       = {https://proceedings.mlr.press/v235/wang24z.html},
  bibsource = {dblp computer science bibliography, https://dblp.org}
}

@article{DBLP:journals/corr/abs-2311-12022,
  author     = {David Rein and
                Betty Li Hou and
                Asa Cooper Stickland and
                Jackson Petty and
                Richard Yuanzhe Pang and
                Julien Dirani and
                Julian Michael and
                Samuel R. Bowman},
  title      = {{GPQA:} {A} Graduate-Level Google-Proof Q{\&}A Benchmark},
  journal    = {CoRR},
  volume     = {abs/2311.12022},
  year       = {2023},
  url        = {https://doi.org/10.48550/arXiv.2311.12022},
  doi        = {10.48550/ARXIV.2311.12022},
  eprinttype = {arXiv},
  eprint     = {2311.12022},
  bibsource  = {dblp computer science bibliography, https://dblp.org}
}

@inproceedings{DBLP:conf/emnlp/ChenYKLWMXWX23,
  author    = {Wenhu Chen and
               Ming Yin and
               Max Ku and
               Pan Lu and
               Yixin Wan and
               Xueguang Ma and
               Jianyu Xu and
               Xinyi Wang and
               Tony Xia},
  editor    = {Houda Bouamor and
               Juan Pino and
               Kalika Bali},
  title     = {TheoremQA: {A} Theorem-driven Question Answering Dataset},
  booktitle = {Proceedings of the 2023 Conference on Empirical Methods in Natural
               Language Processing, {EMNLP} 2023, Singapore, December 6-10, 2023},
  pages     = {7889--7901},
  publisher = {Association for Computational Linguistics},
  year      = {2023},
  url       = {https://doi.org/10.18653/v1/2023.emnlp-main.489},
  doi       = {10.18653/V1/2023.EMNLP-MAIN.489},
  bibsource = {dblp computer science bibliography, https://dblp.org}
}

@article{DBLP:journals/corr/abs-2211-14275,
  author     = {Jonathan Uesato and
                Nate Kushman and
                Ramana Kumar and
                H. Francis Song and
                Noah Y. Siegel and
                Lisa Wang and
                Antonia Creswell and
                Geoffrey Irving and
                Irina Higgins},
  title      = {Solving math word problems with process- and outcome-based feedback},
  journal    = {CoRR},
  volume     = {abs/2211.14275},
  year       = {2022},
  url        = {https://doi.org/10.48550/arXiv.2211.14275},
  doi        = {10.48550/ARXIV.2211.14275},
  eprinttype = {arXiv},
  eprint     = {2211.14275},
  bibsource  = {dblp computer science bibliography, https://dblp.org}
}

@inproceedings{DBLP:conf/iclr/LightmanKBEBLLS24,
  author    = {Hunter Lightman and
               Vineet Kosaraju and
               Yuri Burda and
               Harrison Edwards and
               Bowen Baker and
               Teddy Lee and
               Jan Leike and
               John Schulman and
               Ilya Sutskever and
               Karl Cobbe},
  title     = {Let's Verify Step by Step},
  booktitle = {The Twelfth International Conference on Learning Representations,
               {ICLR} 2024, Vienna, Austria, May 7-11, 2024},
  publisher = {OpenReview.net},
  year      = {2024},
  url       = {https://openreview.net/forum?id=v8L0pN6EOi},
  bibsource = {dblp computer science bibliography, https://dblp.org}
}

@article{DBLP:journals/corr/abs-2207-05221,
  author     = {Saurav Kadavath and
                Tom Conerly and
                Amanda Askell and
                Tom Henighan and
                Dawn Drain and
                Ethan Perez and
                Nicholas Schiefer and
                Zac Hatfield{-}Dodds and
                Nova DasSarma and
                Eli Tran{-}Johnson and
                Scott Johnston and
                Sheer El Showk and
                Andy Jones and
                Nelson Elhage and
                Tristan Hume and
                Anna Chen and
                Yuntao Bai and
                Sam Bowman and
                Stanislav Fort and
                Deep Ganguli and
                Danny Hernandez and
                Josh Jacobson and
                Jackson Kernion and
                Shauna Kravec and
                Liane Lovitt and
                Kamal Ndousse and
                Catherine Olsson and
                Sam Ringer and
                Dario Amodei and
                Tom Brown and
                Jack Clark and
                Nicholas Joseph and
                Ben Mann and
                Sam McCandlish and
                Chris Olah and
                Jared Kaplan},
  title      = {Language Models (Mostly) Know What They Know},
  journal    = {CoRR},
  volume     = {abs/2207.05221},
  year       = {2022},
  url        = {https://doi.org/10.48550/arXiv.2207.05221},
  doi        = {10.48550/ARXIV.2207.05221},
  eprinttype = {arXiv},
  eprint     = {2207.05221},
  bibsource  = {dblp computer science bibliography, https://dblp.org}
}

@inproceedings{DBLP:conf/iclr/XiongHLLFHH24,
  author    = {Miao Xiong and
               Zhiyuan Hu and
               Xinyang Lu and
               Yifei Li and
               Jie Fu and
               Junxian He and
               Bryan Hooi},
  title     = {Can LLMs Express Their Uncertainty? An Empirical Evaluation of Confidence
               Elicitation in LLMs},
  booktitle = {The Twelfth International Conference on Learning Representations,
               {ICLR} 2024, Vienna, Austria, May 7-11, 2024},
  publisher = {OpenReview.net},
  year      = {2024},
  url       = {https://openreview.net/forum?id=gjeQKFxFpZ},
  bibsource = {dblp computer science bibliography, https://dblp.org}
}

@inproceedings{DBLP:conf/emnlp/ManakulLG23,
  author    = {Potsawee Manakul and
               Adian Liusie and
               Mark J. F. Gales},
  editor    = {Houda Bouamor and
               Juan Pino and
               Kalika Bali},
  title     = {SelfCheckGPT: Zero-Resource Black-Box Hallucination Detection for
               Generative Large Language Models},
  booktitle = {Proceedings of the 2023 Conference on Empirical Methods in Natural
               Language Processing, {EMNLP} 2023, Singapore, December 6-10, 2023},
  pages     = {9004--9017},
  publisher = {Association for Computational Linguistics},
  year      = {2023},
  url       = {https://doi.org/10.18653/v1/2023.emnlp-main.557},
  doi       = {10.18653/V1/2023.EMNLP-MAIN.557},
  bibsource = {dblp computer science bibliography, https://dblp.org}
}

@article{DBLP:journals/nature/FarquharKKG24,
  author    = {Sebastian Farquhar and
               Jannik Kossen and
               Lorenz Kuhn and
               Yarin Gal},
  title     = {Detecting hallucinations in large language models using semantic entropy},
  journal   = {Nat.},
  volume    = {630},
  number    = {8017},
  pages     = {625--630},
  year      = {2024},
  url       = {https://doi.org/10.1038/s41586-024-07421-0},
  doi       = {10.1038/S41586-024-07421-0},
  bibsource = {dblp computer science bibliography, https://dblp.org}
}

@inproceedings{ZhangEtAl2022CBLUE,
  author    = {Ningyu Zhang and
               Mosha Chen and
               Zhen Bi and
               Xiaozhuan Liang and
               Lei Li and
               Xin Shang and
               Kangping Yin and
               Chuanqi Tan and
               Jian Xu and
               Fei Huang and
               Luo Si and
               Yuan Ni and
               Guotong Xie and
               Zhifang Sui and
               Baobao Chang and
               Hui Zong and
               Zheng Yuan and
               Linfeng Li and
               Jun Yan and
               Hongying Zan and
               Kunli Zhang and
               Buzhou Tang and
               Qingcai Chen},
  title     = {{{CBLUE}}: A Chinese Biomedical Language Understanding Evaluation Benchmark},
  booktitle = {Proceedings of the 60th Annual Meeting of the Association for
               Computational Linguistics (Volume 1: Long Papers)},
  pages     = {7888--7915},
  publisher = {Association for Computational Linguistics},
  year      = {2022},
  url       = {https://aclanthology.org/2022.acl-long.544/},
  doi       = {10.18653/v1/2022.acl-long.544}
}

@inproceedings{ZhangEtAl2022OntoProtein,
  author    = {Ningyu Zhang and
               Zhen Bi and
               Xiaozhuan Liang and
               Siyuan Cheng and
               Haosen Hong and
               Shumin Deng and
               Jiazhang Lian and
               Qiang Zhang and
               Huajun Chen},
  title     = {{OntoProtein}: Protein Pretraining With Gene Ontology Embedding},
  booktitle = {The Tenth International Conference on Learning Representations},
  year      = {2022},
  url       = {https://openreview.net/forum?id=yfe1VMYAXa4}
}

@inproceedings{BiEtAl2024OceanGPT,
  author    = {Zhen Bi and
               Ningyu Zhang and
               Yida Xue and
               Yixin Ou and
               Daxiong Ji and
               Guozhou Zheng and
               Huajun Chen},
  title     = {{OceanGPT}: A Large Language Model for Ocean Science Tasks},
  booktitle = {Proceedings of the 62nd Annual Meeting of the Association for
               Computational Linguistics (Volume 1: Long Papers)},
  pages     = {3357--3372},
  publisher = {Association for Computational Linguistics},
  year      = {2024},
  url       = {https://aclanthology.org/2024.acl-long.184/},
  doi       = {10.18653/v1/2024.acl-long.184}
}

@article{HuEtAl2025HealthcareLLM,
  author  = {Zhenlin Hu and
             Zhizhi Peng and
             Zhen Bi and
             Qing Shen and
             Zhenfang Liu and
             Jungang Lou and
             Xin Luo},
  title   = {Advancing Healthcare With Large Language Models: Techniques and Application},
  journal = {IEEE/CAA Journal of Automatica Sinica},
  volume  = {12},
  number  = {12},
  pages   = {2371--2398},
  year    = {2025},
  url     = {https://doi.org/10.1109/JAS.2025.125540},
  doi     = {10.1109/JAS.2025.125540}
}

@article{WeiEtAl2025AgenticScience,
  author     = {Jiaqi Wei and
                Yuejin Yang and
                Xiang Zhang and
                Yuhan Chen and
                Xiang Zhuang and
                Zhangyang Gao and
                Dongzhan Zhou and
                Guangshuai Wang and
                Zhiqiang Gao and
                Juntai Cao and
                Zijie Qiu and
                Xuming He and
                Qiang Zhang and
                Chenyu You and
                Shuangjia Zheng and
                Ning Ding and
                Wanli Ouyang and
                Nanqing Dong and
                Yu Cheng and
                Siqi Sun and
                Lei Bai and
                Bowen Zhou},
  title      = {From {AI} for Science to Agentic Science: A Survey on Autonomous
                Scientific Discovery},
  journal    = {CoRR},
  volume     = {abs/2508.14111},
  year       = {2025},
  url        = {https://arxiv.org/abs/2508.14111},
  eprinttype = {arXiv},
  eprint     = {2508.14111}
}

\section*{Additional Definition}
\subsection*{Scientific Conditional Memory Benefit}
Existing evaluation of conditional memory typically relies on overall
performance improvement. However, this does not distinguish gains on cases
where the base model approaches its scientific knowledge boundary from gains
on cases that it already solves. We therefore separately define the
boundary-set gain,
\begin{equation}
\Delta S_{\mathcal{B}} = S(M_{\mathrm{mem}}, \mathcal{B}) - S(M_{\mathrm{base}}, \mathcal{B})
\end{equation}
and the non-boundary-set gain,
\begin{equation}
\Delta S_{\overline{\mathcal{B}}} = S(M_{\mathrm{mem}}, \overline{\mathcal{B}}) - S(M_{\mathrm{base}}, \overline{\mathcal{B}})
\end{equation}
in addition to the aggregate gain
\begin{equation}
    \Delta S = S(M_{\mathrm{mem}}, \mathcal{D}) - S(M_{\mathrm{base}}, \mathcal{D})
\end{equation}
over the complete evaluation set. The \textbf{Scientific Conditional Memory
Benefit (SCMB)} is
\begin{equation}
    S_{CMB} = \Delta S_{\mathcal{B}} - \Delta S_{\overline{\mathcal{B}}}
\end{equation}
and measures whether memory provides larger gains on scientific-boundary
cases than on non-boundary cases.

\section*{Experimental Details and Reproducibility}
\subsection*{SFT Data Synthesis}
\paragraph{BioProBench.}
We synthesize protocol-reasoning SFT data by first clustering the
BioProBench protocol collection to identify frequently occurring
knowledge points. We associate these knowledge points with the original
protocol collection and retain the original protocols in the corresponding
non-isolated clusters as the source pool, rather than sampling protocols
uniformly. To prevent data leakage, every source-protocol step that appears in
the test data is marked as reserved and is not used to construct a new
question.

For each selected source protocol, we generate Error Correction (ERR), Step
Ordering (ORD), and Protocol Question Answering (PQA) queries with task-type
and ordering-length distributions aligned with the benchmark. A
protocol-grounded reviewer filters every query for template compliance,
factual consistency with the source protocol, and answer correctness. For each
accepted query, an API-based generator produces a \texttt{<think>} rationale
conditioned on the full source protocol and the known answer; a second review
checks the response format, biological reasoning, and final answer. A final
deterministic check extracts the answer tag and compares it with the stored
answer. The resulting validated examples are combined with the existing
ERR/ORD records to form the Bio SFT set.

\paragraph{ChemCoTBench.}
We use each original ChemCoTDataset query as the instruction and obtain the
target answer from its structured metadata. When a record provides a
natural-language \texttt{raw\_cot}, we first evaluate its format, chemical
reasoning quality, and consistency with the target. High-quality rationales
are retained and normalized into the standard \texttt{<think>} response
format. Only when \texttt{raw\_cot} is missing or fails this quality gate do
we use the corresponding \texttt{struct\_cot} as a reasoning skeleton and ask
the generator to expand it into a natural-language response. The expansion is
conditioned on the query, the complete structured rationale, and the target
answer, so that it elaborates the chemical reasoning without changing the
target.

All Chem responses undergo format and reasoning review. For molecule-valued
answers, we additionally canonicalize generated and reference SMILES with
RDKit before exact comparison; other answer types are directly compared with
their ground truth. The final Chem SFT set combines existing non-reaction
examples with validated additions.

\subsection*{Compute Environment}
Tables~\ref{tab:compute-hardware} and \ref{tab:compute-software} record the
hardware, system, and principal software versions used for SFT and local
generation. The BioProBench and ChemCoTBench execution environments
follow their respective original benchmark setups. Each SFT and generation job
used one GPU. CPU preprocessing and metric computation are not included in
any latency or GPU-memory comparison. The released environment files provide
complete dependency records for SFT and local generation.

\begin{table*}[t]
\centering
{\small
\setlength{\tabcolsep}{4pt}
\begin{tabular}{>{\raggedright\arraybackslash}p{0.17\textwidth}>{\raggedright\arraybackslash}p{0.37\textwidth}>{\raggedright\arraybackslash}p{0.37\textwidth}}
\toprule
\textbf{Component} & \textbf{SFT} & \textbf{Local benchmark generation} \\
\midrule
Accelerator & 1$\times$ NVIDIA A800-SXM4-80GB (81,920\,MiB) & 1$\times$ NVIDIA GeForce RTX 4090 D (24,564\,MiB) \\
Job allocation & 16 Slurm CPUs; 128\,GB host memory & 16 Slurm CPUs; 128\,GB host memory \\
Host CPU & 2$\times$ Intel Xeon Platinum 8358P at 2.60\,GHz (64 physical cores; 128 hardware threads) & 2$\times$ Intel Xeon Gold 6430 (64 physical cores; 128 hardware threads) \\
Installed host memory & 1,000,000\,MB & 490,000\,MB \\
Operating system & Rocky Linux 9.5; kernel \texttt{5.14.0-503.40.1.el9\_5.x86\_64} & Rocky Linux 9.7; kernel \texttt{5.14.0-611.16.1.el9\_7.x86\_64} \\
GPU driver / reported CUDA & 550.144.03 / 12.4 & 580.95.05 / 13.0 \\
\bottomrule
\end{tabular}
}
\caption{Recorded hardware and system environment. CPU and memory values in
the job-allocation row are the Slurm resources allocated to each single-GPU
job; installed host memory is reported separately.}
\label{tab:compute-hardware}
\end{table*}

\begin{table*}[t]
\centering
{\small
\setlength{\tabcolsep}{4pt}
\begin{tabular}{>{\raggedright\arraybackslash}p{0.24\textwidth}>{\raggedright\arraybackslash}p{0.68\textwidth}}
\toprule
\textbf{Software group} & \textbf{SFT and local generation} \\
\midrule
Runtime & Python 3.12.13 \\
Model and generation stack & PyTorch 2.6.0; Transformers 5.8.0; PEFT 0.19.1; Engram-PEFT 1.2.6 \\
Training and data stack & TRL 1.3.0; Accelerate 1.13.0; Datasets 4.8.5; Tokenizers 0.22.2 \\
Numerical stack & NumPy 2.4.4 \\
\bottomrule
\end{tabular}
}
\caption{Principal recorded software versions for SFT and local generation.}
\label{tab:compute-software}
\end{table*}

\begin{table*}[t]
\centering
{\footnotesize
\setlength{\tabcolsep}{5pt}
\renewcommand{\arraystretch}{0.75}
\begin{tabular}{lll}
\toprule
Setting & BioProBench SFT & ChemCoTBench SFT \\
\midrule
Backbones & \multicolumn{2}{l}{Qwen2.5-7B-Instruct; Qwen3-8B} \\
Epochs & 3 & 3 \\
Maximum sequence length & 4096 & 6144 \\
Per-device batch size & 4 & 3 \\
Gradient accumulation & 4 & 4 \\
Effective single-GPU batch size & 16 & 12 \\
Optimizer & \multicolumn{2}{l}{Adam (LoRA/dense); SparseAdam (hash embeddings)} \\
Adam $\beta$ / $\epsilon$ & \multicolumn{2}{l}{$(0.9,0.999)$ / $10^{-8}$} \\
Learning rates & \multicolumn{2}{l}{LoRA/dense $10^{-5}$; sparse memory $5\times10^{-5}$} \\
Scheduler / warmup & \multicolumn{2}{l}{cosine / first 10\% of update steps} \\
Weight decay / gradient clipping & \multicolumn{2}{l}{0 / max norm 1.0} \\
Precision / checkpointing & \multicolumn{2}{l}{bfloat16 / enabled} \\
Training seed & \multicolumn{2}{l}{42 (Transformers default; no override)} \\
Checkpoint / logging interval & \multicolumn{2}{l}{500 / 10 update steps} \\
LoRA & \multicolumn{2}{l}{$r=8$, $\alpha=16$, dropout 0, no bias, all linear layers} \\
\bottomrule
\end{tabular}
}
\caption{Final SFT hyperparameters. Both backbone families use the same values
unless a model-specific layer is shown.}
\label{tab:sft-hparams}
\end{table*}

\subsection*{Repeated Runs and Randomness}
Every reported task-performance result is computed from three
stochastic generation-and-evaluation runs using the same trained checkpoint,
test split, prompt, and selected routing configuration. Tables report the
arithmetic mean and standard deviation over these three runs, and paired
comparisons use matching runs. The batch-generation utilities may retain
additional run-indexed outputs, but these are not included in the reported
summaries. The random-router control uses 100 fixed masks, each matched to the
proposed router's memory-activation rate.

The SFT entry points use the Transformers training seed of 42, and Engram hash
construction uses seed 0. The released scripts retain run-indexed outputs and
the configurations used for each reported result, enabling the reported
summaries to be re-evaluated with the same checkpoint, test split, prompt, and
routing configuration.

\subsection*{Formal Evaluation Metrics}
\paragraph{Evaluation protocol.}
We use the benchmark-provided parsers and task-specific evaluators. Parsing or
molecular-validity coverage is reported separately so that a high conditional
task score cannot hide a low rate of usable outputs. Let $N$ be the relevant
evaluation-set size and $\mathbf{1}[\cdot]$ the indicator function.

\paragraph{BioProBench Protocol Question Answering (PQA).}
Let $\widehat{y}_i$ and $y_i$ be the predicted and reference answers, let
$z_i=\mathbf{1}[\widehat{y}_i=y_i]$, and let $p_i\in[0,1]$ be the model's
parsed confidence after dividing the reported percentage by 100. We compute
\begin{equation}
\mathrm{Acc}=\frac{1}{N}\sum_{i=1}^{N}z_i,\qquad
\mathrm{Brier}=\frac{1}{N}\sum_{i=1}^{N}(p_i-z_i)^2.
\end{equation}
Accuracy measures exact answer correctness, whereas the Brier score jointly
penalizes incorrect answers and miscalibrated confidence; lower Brier is
better.

\paragraph{BioProBench Step Ordering (ORD).}
For predicted and reference permutations $\widehat{\pi}_i$ and $\pi_i$,
exact match is
\begin{equation}
\mathrm{EM}=\frac{1}{N}\sum_{i=1}^{N}
\mathbf{1}[\widehat{\pi}_i=\pi_i].
\end{equation}
For an instance containing $m_i$ steps, let $C_i$ and $D_i$ be the numbers of
concordant and discordant step pairs. The evaluator pools these pairs across
instances and computes
$\tau=\sum_i(C_i-D_i)/\sum_i(m_i(m_i-1)/2)$. EM tests recovery of the complete
protocol order, while Kendall's $\tau$ gives partial credit for pairwise
ordering agreement.

\paragraph{BioProBench Error Correction (ERR) and our balanced extension.}
Accuracy is the fraction of correctly predicted Boolean labels. For each
class $c\in\{\mathrm{False},\mathrm{True}\}$, define
\begin{equation}
\begin{array}{rcl}
P_c&=&\displaystyle\frac{\mathrm{TP}_c}
{\mathrm{TP}_c+\mathrm{FP}_c},\qquad
R_c=\displaystyle\frac{\mathrm{TP}_c}
{\mathrm{TP}_c+\mathrm{FN}_c},\\[8pt]
F1_c&=&\displaystyle\frac{2P_cR_c}{P_c+R_c},
\end{array}
\end{equation}
with a zero value when the corresponding denominator is zero. The official
BioProBench evaluator treats \texttt{False} (the protocol step is
incorrect, i.e., an error is present) as the sole positive class. We retain
its parsing and accuracy calculation but report
\begin{equation}
\begin{array}{rcl}
\mathrm{Macro\mbox{-}P}
&=&(P_{\mathrm{False}}+P_{\mathrm{True}})/2,\\
\mathrm{Macro\mbox{-}R}
&=&(R_{\mathrm{False}}+R_{\mathrm{True}})/2,\\
\mathrm{Macro\mbox{-}F1}
&=&(F1_{\mathrm{False}}+F1_{\mathrm{True}})/2.
\end{array}
\end{equation}
This balanced extension evaluates both error detection and correct-step
recognition with equal class weight, reducing sensitivity to the arbitrary
choice of a single positive class and to class imbalance. Parser failures are
counted and reported separately from these class metrics.

\paragraph{ChemCoTBench molecular understanding.}
For functional-group and ring-count subtasks, MAE is
$N_v^{-1}\sum_i|\widehat{n}_i-n_i|$ over the $N_v$ valid, parsed outputs.
For categorical or equivalence subtasks, accuracy is the fraction of exact
task decisions. For Murcko-scaffold extraction, TMS is the mean Tanimoto
similarity between the predicted and reference scaffold fingerprints:
\begin{equation}
\mathrm{TMS}(A,B)=\frac{|A\cap B|}{|A\cup B|}.
\end{equation}
Following the evaluator, exactly matching scaffolds receive 1; otherwise,
Tanimoto similarity is computed from radius-2, 1024-bit Morgan fingerprints
of the Murcko scaffolds. These metrics respectively measure numerical
counting error, discrete structural understanding, and graded scaffold
similarity.

\paragraph{ChemCoTBench molecular editing.}
Editing accuracy is the number of outputs that carry out the requested
addition, deletion, or substitution divided by the total number of examples;
unextractable or invalid molecules count as failures. Validity is the fraction
of outputs from which the evaluator can extract a target molecule that RDKit
accepts as a valid molecule. Accuracy measures instruction compliance, while
validity detects chemically unusable generations.

\paragraph{ChemCoTBench molecular optimization.}
For target property oracle $f$ and generated molecule
$\widehat{m}_i$, the improvement is
$\Delta_i=f(\widehat{m}_i)-f(m_i)$. Unextractable, invalid, or unscorable
outputs receive $\Delta_i=0$. Mean improvement is the evaluator's
5th--95th-percentile winsorized mean of $\{\Delta_i\}$, and success rate is
$N^{-1}\sum_i\mathbf{1}[\Delta_i>0]$. Validity is the fraction of generated
molecules that are valid and receive a property score; extraction rate is the
fraction of responses from which the final target SMILES can be extracted.
Together these metrics separate the magnitude and frequency of property
improvement from output usability.

\paragraph{Validity convention.}
For every ChemCoTBench subtask, \emph{Val.} is a coverage measure: the fraction
of all examples whose outputs can be parsed and admitted to that subtask's
evaluation path. Consequently, MolUnd reports separate validity values for
the MAE, accuracy, and TMS subtask groups rather than sharing one denominator.

\subsection*{Final Hyperparameters}
\paragraph{SFT configuration.}
Table~\ref{tab:sft-hparams} consolidates the final training settings. SFT is
performed once per backbone--domain pair; the three reported runs are
generation repeats from the resulting fixed checkpoint. Both domain-specific
SFT sets contain tens of thousands of examples, all of which are used for
training.

\paragraph{Memory adapter.}
For Qwen2.5, memory is injected at zero-indexed layers $[1,14]$; for Qwen3,
at $[1,18]$. Both use embedding dimension 1024, n-grams $[2,3]$,
1,131,200 hash buckets per n-gram size, eight hash heads per n-gram,
tokenizer compression, hash multiplier 4, summed multi-head features, a
kernel-4 convolution with dilation 3, zero-initialized convolution and gate,
sparse embeddings, Engram learning-rate multiplier 5, Engram weight decay 0,
and hash seed 0. For Memory+LoRA, the existing LoRA parameters remain
trainable while the memory adapter is trained. The Qwen2.5 composite has
347,059,200 trainable parameters (20,185,088 LoRA and 326,874,112 memory)
out of 7,962,675,712 total parameters. The Qwen3 composite has 353,969,152
trainable parameters (21,823,488 LoRA and 332,145,664 memory) out of
8,544,704,512 total parameters. The observed final-job SFT wall times were
30:47:14 (Qwen2.5 Bio), 36:25:23 (Qwen3 Bio), 27:36:31 (Qwen2.5 Chem), and
33:18:07 (Qwen3 Chem).

\paragraph{Generation.}
Local generation uses the thinking-enabled chat template, a maximum of 4096
new tokens, and one sampled sequence per input. BioProBench uses
bfloat16, temperature 0.6, top-$p$ 0.95, and top-$k$ 50. Its Qwen2.5 batch
sizes are 4 for ERR/PQA and 3 for ORD; its Qwen3 batch sizes are 4 for ERR/PQA
and 2 for ORD. ChemCoTBench uses float16, temperature 0.2, and top-$p$ 0.8,
with batch size 8 for Qwen2.5 and 5 for Qwen3. KV caching is enabled during
generation. Prompt templates, answer-parsing rules, task-specific schemas, and
run-indexed output paths are included in the released code.

External and internal routing is empirical, based on knowledge-boundary and
knowledge-circuit theory and existing reasoning samples; hyperparameter search
selects its parameters, which are provided in the released code.

\begin{figure*}[t]
    \centering
    \includegraphics[width=\textwidth]{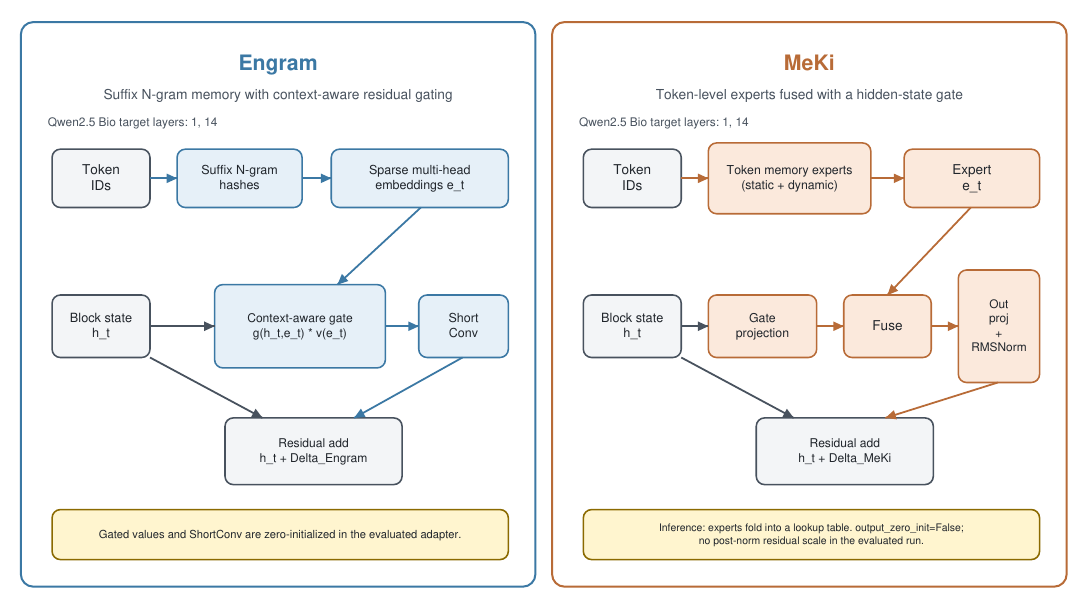}
    \caption{Structural comparison of the Engram and MeKi memory adapters in
    the evaluated Qwen2.5 BioProBench configuration. Engram combines
    suffix n-gram retrieval with context-aware gating and a short-convolution
    residual path. MeKi combines token-level expert lookup with hidden-state
    gating, fusion, output projection, and RMS normalization before direct
    residual addition. The lower MeKi callout records configuration details of
    the diagnostic run, including \texttt{output\_zero\_init=False}; it is not
    a causal attribution of the generation-collapse result.}
    \label{fig:meki-engram-architecture}
\end{figure*}

\section*{MeKi and Engram Adapter Architectures}
Figure~\ref{fig:meki-engram-architecture} compares the concrete memory paths
used in the diagnostic experiment. Engram retrieves suffix n-gram embeddings,
uses the current hidden state to contextually gate their contribution, and
passes the gated residual through a short convolution before injection. The
evaluated MeKi configuration retrieves token-level experts, fuses them with a
hidden-state gate, applies an output projection and RMS normalization, and then
adds the resulting residual to the target module output. Its training-time
static and dynamic expert features are folded into a lookup table at inference.
Both Qwen2.5 diagnostic configurations target layers 1 and 14. The comparison
describes the evaluated implementations and does not attribute the observed
MeKi generation failure to any single architectural component.



\end{document}